\documentclass[10pt,twocolumn,letterpaper]{article}

\usepackage[pagenumbers]{cvpr} 

\newcommand{\nats}{\mathbb{N}}

\newcommand{\reals}{\mathbb{R}}
\newcommand{\bx}{\mathbf{x}}

\newcommand{\bM}{\mathbf{M}}
\newcommand{\calB}{\mathcal{B}}
\newcommand{\cyc}[1]{\langle #1\rangle}

\usepackage{svg}
\usepackage{todonotes}
\usepackage{relsize}
\usepackage{tabularx}

\newcommand{\smallsec}[1]{\vspace{2pt} \noindent \textbf{#1}.}
\newcommand{\metric}{Attention-IoU}

\newcommand{\apnorm}{$\mathrm{AP_N}$}

\usepackage[accsupp]{axessibility}
\usepackage{nowidow}

\definecolor{cvprblue}{rgb}{0.21,0.49,0.74}
\usepackage[pagebackref,breaklinks,colorlinks,allcolors=cvprblue]{hyperref}

\def\paperID{12014} 
\def\confName{CVPR}
\def\confYear{2025}

\title{Attention IoU: Examining Biases in CelebA using Attention Maps}

\author{Aaron Serianni\qquad Tyler Zhu \qquad Olga Russakovsky \qquad Vikram V. Ramaswamy\\
Princeton University\\
{\tt\small \{serianni, tylerzhu, olgarus, vr23\}@princeton.edu}
}

\begin{document}
\maketitle
\begin{abstract}
    Computer vision models have been shown to exhibit and amplify biases across a wide array of datasets and tasks. Existing methods for quantifying bias in classification models primarily focus on dataset distribution and model performance on subgroups, overlooking the internal workings of a model. We introduce the \metric{} (Attention Intersection over Union) metric and related scores, which use attention maps to reveal biases within a model's internal representations and identify image features potentially causing the biases. First, we validate \metric{} on the synthetic Waterbirds dataset, showing that the metric accurately measures model bias. We then analyze the CelebA dataset, finding that \metric{} uncovers correlations beyond accuracy disparities. Through an investigation of individual attributes through the protected attribute of \texttt{Male}, we examine the distinct ways biases are represented in CelebA. Lastly, by subsampling the training set to change attribute correlations, we demonstrate that \metric{} reveals potential confounding variables not present in dataset labels. Our code is available at \url{https://github.com/aaronserianni/attention-iou}.

\end{abstract}
\section{Introduction}
\label{sec:intro}
\looseness=-1
Biases in computer vision models can lead to failures in model performance and unequal behavior for different groups. These biases are often caused by spurious correlations, where a model relies on an attribute that is associated with, but not causally related to, the target. A model dependent on such spurious correlations might then perform poorly on out-of-distribution test data or exhibit low accuracy for groups for which the correlation does not hold. For example, models have been shown to be biased towards low-level features such as texture and image spectra~\cite{gavrikov_can_2024,xiao_noise_2021,geirhos_imagenet-trained_2019}, and high-level attributes including background and contextual objects~\cite{singh_dont_2020}. This becomes more concerning for tasks involving people, since these correlations can cause models to discriminate against societally protected groups such as gender, race, age, ethnicity, and income~\cite{buolamwini_gender_2018,zhao_men_2017,zhao_understanding_2021,burns_women_2019,devries2019objectrecognition,shankar_no_2017}. 

\begin{figure}
    \centering
    \begin{subfigure}{0.24\linewidth}
        \centering
        \includegraphics[width=\linewidth]{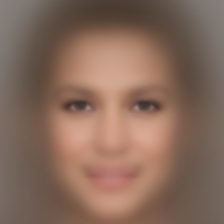}
        \caption*{Average Image}
    \end{subfigure}
    \begin{subfigure}{0.24\linewidth}
        \centering
        \includegraphics[width=\linewidth]{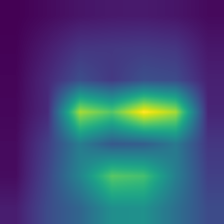}
        \caption*{\texttt{Male}}
    \end{subfigure}
    \begin{subfigure}{0.24\linewidth}
        \centering
        \includegraphics[width=\linewidth]{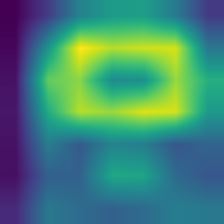}
        \caption*{\texttt{Blond\_Hair}}
    \end{subfigure}
    \begin{subfigure}{0.24\linewidth}
        \centering
        \includegraphics[width=\linewidth]{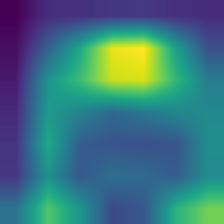}
        \caption*{\texttt{Wavy\_Hair}}
    \end{subfigure}
    \\
    \begin{subfigure}{\linewidth}
        \centering
        \includegraphics[width=\linewidth]{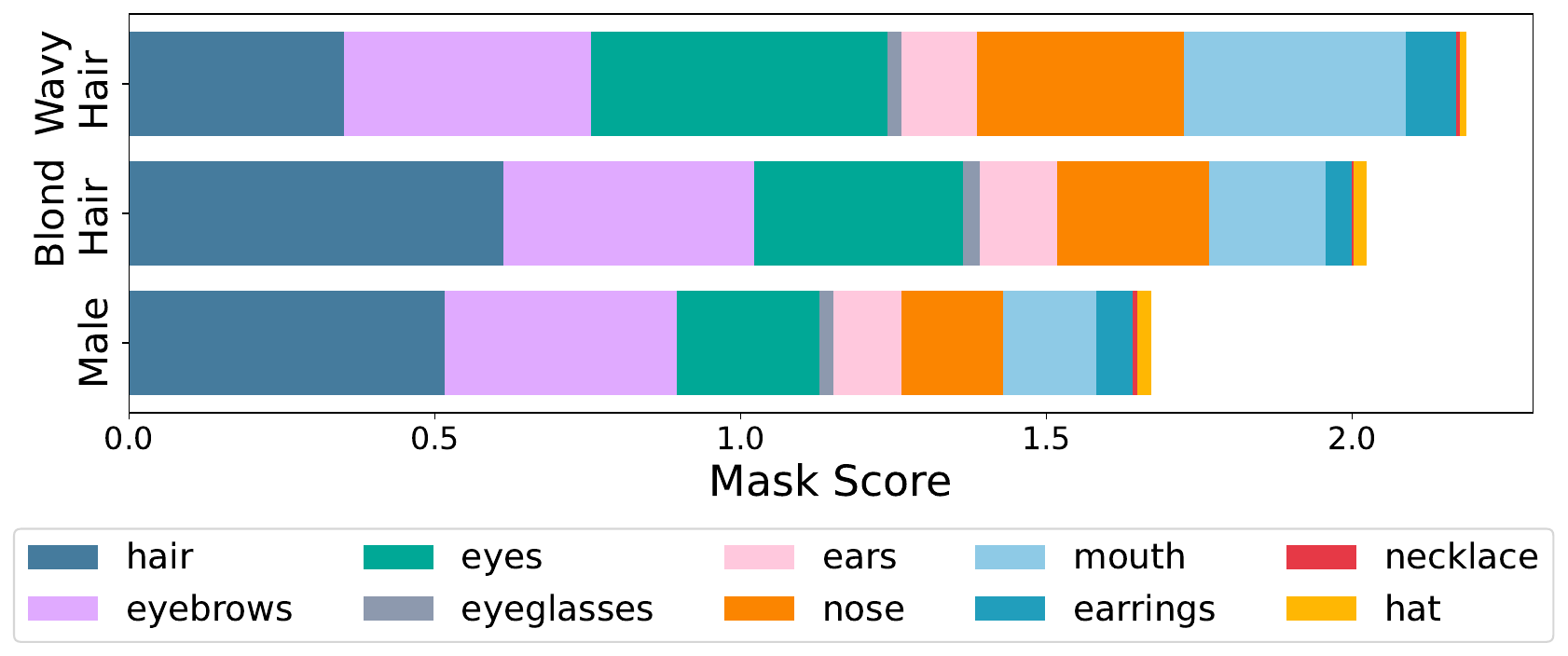}
    \end{subfigure}
    \caption{We use attention maps to understand which image regions a model relies on for the target classification task. Our proposed \metric{} framework provides insights into how models represents biases between correlated attributes. For example, consider the spatially related attributes of blond and wavy hair in the CelebA dataset~\cite{liu_deep_2015}, which have similar label correlations to the \texttt{Male} label. They are attended to differently by the model, with blond hair appearing closer to \texttt{Male} in both average attention map (\emph{top row}) and the \metric{} mask score (\emph{bottom row}). 
    Thus, \metric{} reveals that blond hair, when compared to wavy hair, has a  spurious correlation with \texttt{Male} that is not present in the dataset labels.
    }
    \label{fig:teaser}
\end{figure}

Past works have extensively investigated biases and spurious correlations through the lens of dataset labeling and model accuracy. For example, fairness metrics reveal disparities in model accuracy between groups or individuals (see \cite{caton_fairness_2024,verma_fairness_2018,mehrabi_survey_2022,pessach_review_2023} for surveys). Others have created tools to surface biases by analyzing and categorizing objects, gender, skin tone, geographical labels, among others, sometimes in combination with model predictions and unsupervised techniques~\cite{bellamy_ai_2019,wang_revise_2022,krishnakumar_udis_2021}. Many studies have also explored methods to mitigate the effects of spurious correlations in datasets~\cite{zhao_men_2017,burns_women_2019,wang_towards_2020,parraga_fairness_2023,izmailov_feature_2024}.

These approaches to the discovery and measurement of spurious correlations using dataset inputs and model outputs have revealed many biases exhibited by computer vision models. However, they are often only able to find biases at a coarse level, restricted to the binary labels present within the dataset. For example, while these metrics excel at identifying when the classification of a person's attributes might depend on gender, they are unable to highlight the specific features of the person's gender presentation that the model uses to make a prediction. 
In the absence of fine-grained labels, interpretability methods hold the potential to reveal representations of correlations within a model, and how they might affect the model's output.


In this paper, we propose \textit{\metric{}}, a generalized intersection-over-union metric that uses attention maps to measure biases in image classification models. We specifically aim to quantify spurious correlations for when a model relies on regions of images that are not directly relevant to the target classification tasks. For example, within the CelebA dataset~\cite{liu_deep_2015, lee_maskgan_2020} of faces, blond hair is correlated with a person being labeled not male. As such, a model trained to identify the `blond hair' attribute may use gendered aspects faces in addition to using hair features to compute its output. Thus, the model may attend to regions such as the eyes, nose, and mouth as well as hair~(\cref{fig:teaser}). As part of \metric{}, we present two scores: the mask score, where an attention map is compared to a ground-truth feature mask; and the heatmap score, where the attention maps for two different attributes are compared with each other.

We first validate \metric{} on the synthetic Waterbirds~\cite{sagawa_distributionally_2020} dataset, showing that it accurately reflects the bias within the dataset. We then examine CelebA~\cite{liu_deep_2015}, as the dataset is a widely-used benchmark for fairness methods, spanning dataset bias identification to model debiasing, with the \texttt{Male} attribute used as the sensitive attribute.


Through this analysis of CelebA, we demonstrate that \metric{} can identify specific ways in which the protected \texttt{Male} attribute might influence other attributes. We show that attributes can be unevenly influenced by the classifier's representation of the protected \texttt{Male} attribute, and that certain attributes have biases beyond simple correlations in dataset labels. These insights reveal different ways in which computer vision models might be biased, allowing the community to develop better debiasing techniques.


\section{Related Work}
\label{sec:related}

\smallsec{Bias in computer vision} Computer vision models and datasets have been extensively shown to exhibit biases across a wide range of tasks~\cite{meister_gender_2023, fabbrizzi_survey_2022, torralba_unbiased_2011, wang_balanced_2019, van_miltenburg_stereotyping_2016, hirota_gender_2022, burns_women_2019, wang_are_2021}. 
Models can even amplify disparities from the datasets on which they are trained~\cite{zhao_men_2017, seshadri_bias_2023, caliskan_semantics_2017, zhao_understanding_2021}. 
When biased datasets and models involve people and society, there are significant fairness and societal implications, as models often perform anomalously regarding protected classes including race, gender, and age~\cite{buolamwini_gender_2018, bianchi_easily_2023, luccioni_stable_2023, pahl_female_2022, wolfe_american_2022}.

Past works about identifying biases in computer vision focus either on quantifying bias in the dataset, the output of the trained model, or a combination thereof~\cite{buolamwini_gender_2018, pahl_female_2022, harrison_run_2023}. 
Bias is often quantified by analyzing the distribution of attributes within a dataset, and identifying which attributes have unequal distributions or are underrepresented compared to real-world demographics~\cite{buolamwini_gender_2018, shankar_no_2017}. 
For unannotated attributes, this can be revealed through the use of both image generative models to balance the distribution~\cite{balakrishnan_towards_2020,liang_benchmarking_2023,denton_image_2020}, and vision language models for the fine-grained identification of attributes and unlabeled biases \cite{li_discover_2022,kabra_gelda_2023, kim_discovering_2024}. Other approaches find correlations between labeled attributes and features in the images themselves, such as co-occurring objects~\cite{singh_dont_2020}, stereotypical and offensive portrayals~\cite{bianchi_easily_2023}, or low-level features like pose and color~\cite{meister_gender_2023}. In a trained model, bias identification  is primarily restricted to looking at the model's outputs, often including calculating the accuracy and error rates for various labeled groups within the dataset~\cite{verma_fairness_2018, dixon_measuring_2018, chouldechova_fair_2017, hardt_equality_2016}, or, if groups are unlabeled, using unsupervised techniques to find them~\cite{li2022discover,sohoni2020no,nam_learning_2020,krishnakumar_udis_2021}.

\smallsec{Interpretability methods and metrics} Interpretability for machine learning aims to explain the external behavior of models and give insight into their internal mechanisms. 
Instance or local explanations are the most common interpretability technique for computer vision, describing how the model behaves locally around features in a specific input. The output of this technique is an attention or saliency heatmap, highlighting the areas of the image most responsible for the model's output~\cite{fong2019extremal,petsiuk2018rise,selvaraju_grad-cam_2020,shitole2021sag,simonyan_deep_2014,zeiler_visualizing_2014,zhou_learning_2016,goyal2019counterfactual,vandenhende2022counterfactual,wang2020scout,ribeiro2016lime}. Class activation maps (CAM)~\cite{zhou_learning_2016} and its derivatives, including GradCAM~\cite{selvaraju_grad-cam_2020}, are the most common methods for creating attention maps. 

Attention maps are frequently used qualitatively to evaluate debiasing methods~\cite{tartaglione_end_2021,xu_vitae_2021, singla_understanding_2021, huang_gradient_2023}, or highlight biases in models~\cite{selvaraju_grad-cam_2020}, as Wolfe \etal perform through average heatmaps ~\cite{wolfe_contrastive_2023}.
Krishnakumar \etal and Lee \etal both use attention maps as part of bias visualization systems by highlighting the maps of individual pertinent images~\cite{krishnakumar_udis_2021, lee_viscuit_2022}. Beyond quantitative evaluations, Bang \etal present a method to directly identify model bias using aggregated explanation alignment metrics, focusing on the bias between different model instances~\cite{bang_explanation_2024}.
Some debiasing methods also use attention maps directly, by creating loss functions that integrate attention maps~\cite{singh_dont_2020,rao_studying_2023,li_tell_2018}, or highlighting pertinent image regions through thresholding of the map~\cite{kim_biaswap_2021, pillai_consistent_2022,asgari_masktune_2022}. Specifically, Singh \etal use a loss function to minimize element-wise overlap between the attention maps of an attribute and its co-occurring context, but do not use the maps to evaluate the biases themselves~\cite{singh_dont_2020}.

\section{Method}
\label{sec:method}


Existing bias metrics for computer vision classification models focus on how the models perform with respect to certain groups within a dataset~\cite{hardt_equality_2016, zhao_men_2017}. 
This can involve investigating the distribution of groups in a dataset, differences in accuracy and error rates between groups, or some combination thereof~\cite{verma_fairness_2018}.
These common approaches often only consider the final predictions of models, but in line with other works~\cite{denton_image_2020,balakrishnan_towards_2020,krishnakumar_udis_2021,lee_viscuit_2022}, we aim to understand \emph{why} these biases might occur.
Consider, for example, a dataset where we try to distinguish between waterbirds and landbirds~\cite{sagawa_distributionally_2020}. Here, the birds are correlated with the backgrounds, and most images of waterbirds picture a water background, while most images of landbirds picture a land background. Moreover, assume that a model trained on this dataset struggles to recognize waterbirds pictured on land backgrounds. Metrics that consider the difference in performance between different groups would correctly identify this model as biased. However, we argue that there are multiple forms that this bias could take: 

\begin{figure}
    \centering
    \begin{subfigure}{0.25\linewidth}
        \centering
        \includegraphics[width=\linewidth]{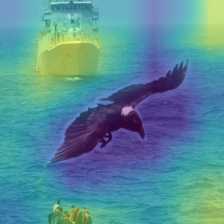}
        \caption*{Background Bias}
    \end{subfigure}
    \quad
    \begin{subfigure}{0.25\linewidth}
        \centering
        \includegraphics[width=\linewidth]{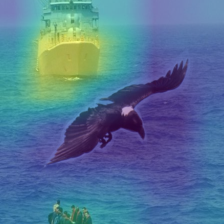}
        \caption*{Object Bias}
    \end{subfigure}
    \quad
    \begin{subfigure}{0.25\linewidth}
        \centering
        \includegraphics[width=\linewidth]{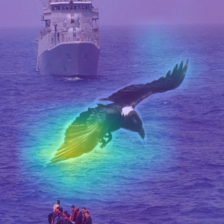}
        \caption*{Depiction Bias}
    \end{subfigure}
    \quad
    \caption{\textbf{Attention maps for a landbird on a water background in the Waterbirds dataset~\cite{sagawa_distributionally_2020}, illustrating possible forms of model bias for incorrect classifications.} (\textit{left}) attending to the whole background; (\textit{center}) attending to a ship instead of the bird; (\textit{right}) only attending to a part of the bird, its wing in flight.}
    \label{fig:biases}
\end{figure}

\begin{itemize}
    \item The model could be using the entire background to identify the bird, and thus, is (incorrectly) using cues from the water background when landbirds are pictured on water (\cref{fig:biases} \textit{left}).
    \item The model could be using specific cues from the background. 
    For example, suppose land backgrounds always contain a tree, while water backgrounds always contain a boat.
    The model could use these cues (rather than the entire background) to classify the image (\cref{fig:biases} \textit{center}). 
    \item Landbirds pictured on water backgrounds could be depicted differently to those pictured on land backgrounds. 
    For example, maybe these birds are pictured mid-flight, making them smaller and thus harder for a model to classify. In this case, the model might be (correctly) using cues from the bird, but the cues learned are not generalized to landbirds on water backgrounds (\cref{fig:biases} \textit{right}). 
\end{itemize}
In order to better understand these differences, we turn to attention maps as a mechanism for revealing which image features are important for the model's decision-making.


The key insight for our bias identification method is the following: 
if a model learns a spurious correlation between a target attribute and a confounding attribute in the dataset, it will learn to use features helpful for the \emph{confounding attribute} instead of the target attribute. 
This lets us quantify bias by comparing a model's attention map for the target attribute to either attention maps of confounding attributes or ground-truth feature maps.


\subsection{GradCAM Preliminaries}
\label{sec:gradcam}
We use Gradient-weighted Class Activation Mapping (GradCAM) to obtain attention maps for target attributes~\cite{selvaraju_grad-cam_2020}. 
Given an input image $\bx$ and target attribute $a$, GradCAM computes the gradient of the class output $y_a$ with respect to the output of a convolutional layer, usually the final layer, to obtain activation maps of the attribute.
A simple gradient-weighted linear combination of the layer's feature activation maps produces the attribute-specific attention map $\mathrm{GradCAM}_a(\bx)$. 
GradCAM was developed for models trained with categorical cross entropy loss, and thus, in its standard implementation, only able create attention maps for positive predictions for a model trained with binary cross entropy loss. For our metric, we instead take the gradient of the absolute value of the class output, $|y_a|$, so that image features that contribute positively to either prediction is attended to in the attention map. For further explanation, see \cref{sec:gradcam-details}.

\subsection{Attention Map Metrics}
\label{sec:method-metric}

Now that we have maps corresponding to attention maps and ground-truth feature masks, all we need is a way to compare the maps. 
The metric should be able to compare two real-valued attention maps with each other, as well as an attention map with a binary ground-truth feature mask.
Two commonly-used metrics for evaluating attention maps, the pointing game~\cite{zhang_top-down_2016} and intersection-over-union (IoU), both fail this requirement as they require a binary mask for one of their inputs. Furthermore, as image attributes can vary drastically in pixel area, such as hair vs. eye color, the metric should be \textit{size invariant} and remain constant if the two maps scale proportionally with each other.

Based on these constraints, we propose a generalized IoU metric, which we refer to as \emph{\metric{}}, that works on weighted dense-pixel maps and is size and scale invariant. 
Given two maps $\bM_1, \bM_2\in \mathbb{R}^{h\times w}$, which can be either attention maps or feature masks, denote their $L_1$ normalized maps as $\widehat{\bM}_i = \bM_i/\|\bM_i\|_1$, which are akin to probability density functions.
The metric is defined as
\begin{align}
    \calB_{\text{A-IoU}}(\bM_1, \bM_2) &= \dfrac{\cyc{\widehat{\bM}_1, \widehat{\bM}_2}_F}{\left\|\frac{\widehat{\bM}_1+\widehat{\bM}_2}{2}\right\|_F^2} 
     = \dfrac{\sum_{i,j} (\widehat{\bM}_1)_{ij}\cdot (\widehat{\bM}_2)_{ij}}{\sum_{ij} \left(\frac{\widehat{\bM}_1 + \widehat{\bM}_2}{2}\right)_{ij}^2}
\end{align}
where $\cyc{\mathbf{A},\mathbf{B}}_F$ is the Frobenius inner product, \ie, the sum of the element-wise matrix product, and $\|\mathbf{A}\|_F^2$ is the Frobenius norm, \ie, the sum of squared entries of the matrix.

The $L_1$ normalization ($\widehat{\bM}_1 = \bM_1/\|\bM_1\|_1$) inside the products makes $\calB_{\text{A-IoU}}$ scale invariant to values of the maps. 
The numerator of our metric calculates a weighted intersection between the two maps. 
If one is binary, then this reports the overall mass focused on relevant mask areas, and when both are continuous, then this simply weights the mass by the corresponding pixel-wise probability.
The denominator of our metric is a union of the two maps. 
We average both maps so that the resulting matrix still has values in $[0,1]$. For full proofs of the invariants, see \cref{sec:proofs}.

This metric has desirable properties similar to IoU; for example, if $\bM_1 = \bM_2$, $\calB_{\text{A-IoU}}(\bM_1, \bM_2)$ is 1, and if the maps are completely disjoint then $\calB_{\text{A-IoU}}$ is 0. 
Since \metric{} allows for continuous scores, if $\bM_1$ and $\bM_2$ overlap, then as the weight in their intersection increases (and decreases, respectively), so does $\calB_{\text{A-IoU}}$. 

\subsection{Bias Scores}
\label{sec:method-scores}
Using \metric{}, we define two methods to score biases in a model for a given target attribute. The first one, the \emph{heatmap score}, compares the attention map for the target attribute $t$ with the attention map of a chosen protected $p$ attribute between each input image. Given a set of images $\{\bx_i\}_{i=1}^n$, the score's formulation is
\begin{align}
    & \ \text{Attention-IoU}_{\mathrm{Heatmap}}(t, p) = \\
    & \qquad\frac{1}{n}\sum_{i=1}^n \calB_{\text{A-IoU}}(\mathrm{GradCAM}_t(\bx_i), \mathrm{GradCAM}_p(\bx_i)). \nonumber
\end{align}
The \textit{mask score} is computed between the target's attention map and a chosen ground-truth feature mask $\mathrm{mask}_f(\bx)$, corresponding to the specific input image. 
As the size of the attention map is the size of the final convolution layer, whereas the feature mask is the size of the input image, the feature mask is downsampled with bilinear interpolation:
\begin{align}
    & \ \text{Attention-IoU}_{\mathrm{Mask}}(t, f) = \\
    & \qquad\frac{1}{n}\sum_{i=1}^n \calB_{\text{A-IoU}}(\mathrm{GradCAM}_t(\bx_i), \mathrm{interp}(\mathrm{mask}_f(\bx_i))). \nonumber
\end{align}

\smallsec{Advantages of \metric{}}
\metric{} has several advantages over existing bias detection methods. First, since the metric is based on attention maps, it highlights specific regions of the sensitive attribute that most contribute to the target attribute prediction. Thus, we are able to identify bias at a more fine-grained level than other bias metrics. Next, by visualizing the scores separately for different types of images, we can infer if the bias is different for the different sets. For example, this allows us to understand if the features of the sensitive attribute are used solely when the attribute takes on a particular value.  
Finally, the metric allows us to unearth potential confounding variables; \ie, when the bias is due to more than the simple proportion of labels within the training dataset. 

One limitation to this attention-based approach is that attention maps only convey spatial information about what the model is attending to in an image. Information regarding shape, color, or texture is not included in an attention map. Thus, if a target and confounding attribute are co-located, but the model is attending to different image features within the region containing both attributes, our metric will still indicate high correlation between the two attributes. Despite this limitation, in the next two sections we show how \metric{} can be used to closely examine a dataset. 

\section{Validating the metric}
\label{sec:validation}

To start, we test the proposed metric on Waterbirds~\cite{sagawa_distributionally_2020}. This simple synthetic dataset is constructed by combining cropped bird images from the CUB dataset~\cite{wah_caltech-ucsd_2011} with backgrounds from the Places dataset~\cite{zhou_places_2018}.  In the dataset birds are labeled as either a waterbird or landbird, and backgrounds are similarly labeled as land or water. The dataset can be constructed with different levels of correlation between the bird and the background, introducing a single axis of bias within the dataset. Moreover, masks of the bird and background are clearly available within this dataset, which can be used to compute \metric{}. 

\smallsec{Experimental setup} 
Following prior work, we place a specified percentage (between 50\%-100\%) of the waterbirds on a water background, with the remaining 0\%-50\% of the waterbirds are placed on a land background, and similarly for landbirds and land backgrounds. The validation and test sets are unbiased with a bird being 50\% likely to align with its background. 
We followed Sagawa \etal~\cite{sagawa_distributionally_2020} in using the official train-test split of the CUB dataset, composed of 5,994 training images and 5,794 testing images, and randomly choosing 20\% of the training images to form the validation set. 
The test set was used to compute the overall accuracy, per-group accuracy, and \metric{}. We used ResNet-18~\cite{he_deep_2016} pretrained on ImageNet~\cite{russakovsky_imagenet_2015} as our model, trained on Waterbirds using categorical cross-entropy loss and an Adam optimizer~\cite{kingma_adam_2017} (learning rate $0.001$, weight decay $0.0001$). 
Input images are rescaled to be $224\times224$, and augmented using random crops and horizontal flips during training. 
Models were trained for 10 epochs, with a batch size of 64. We report averages and standard deviations over 20 individually trained models.

\begin{figure}[t]
   \setlength{\tabcolsep}{-1pt}
   \centering
    \begin{tabular}{c c c c c}
         \includegraphics[width=0.193\linewidth]{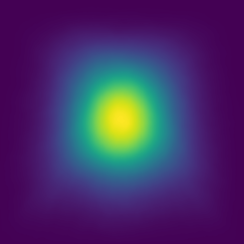} \,& 
         \includegraphics[width=0.193\linewidth]{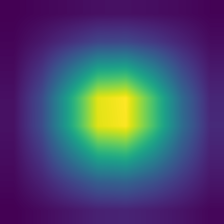} \,& \includegraphics[width=0.193\linewidth]{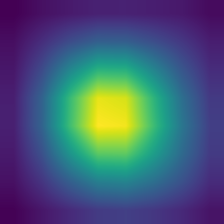} \,& \includegraphics[width=0.195\linewidth]{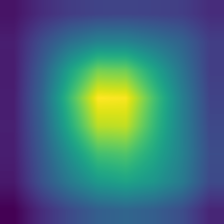}  \,& \includegraphics[width=0.193\linewidth]{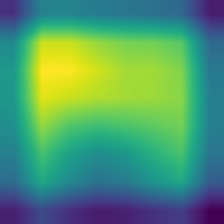} \\
         Bird Mask & 70\% bias & 90\% bias & 95\% bias & 100\% bias \\
    \end{tabular}
    \caption{\textbf{Average bird mask and average heatmaps for Waterbirds at increasing levels of bias.} We see that the model attends less on the bird as the bias increases, as indicated by its mask.}
    \label{fig:waterbirds_heatmaps}
\end{figure}

\begin{figure}[t]
    \centering
    \includegraphics[width=\linewidth]{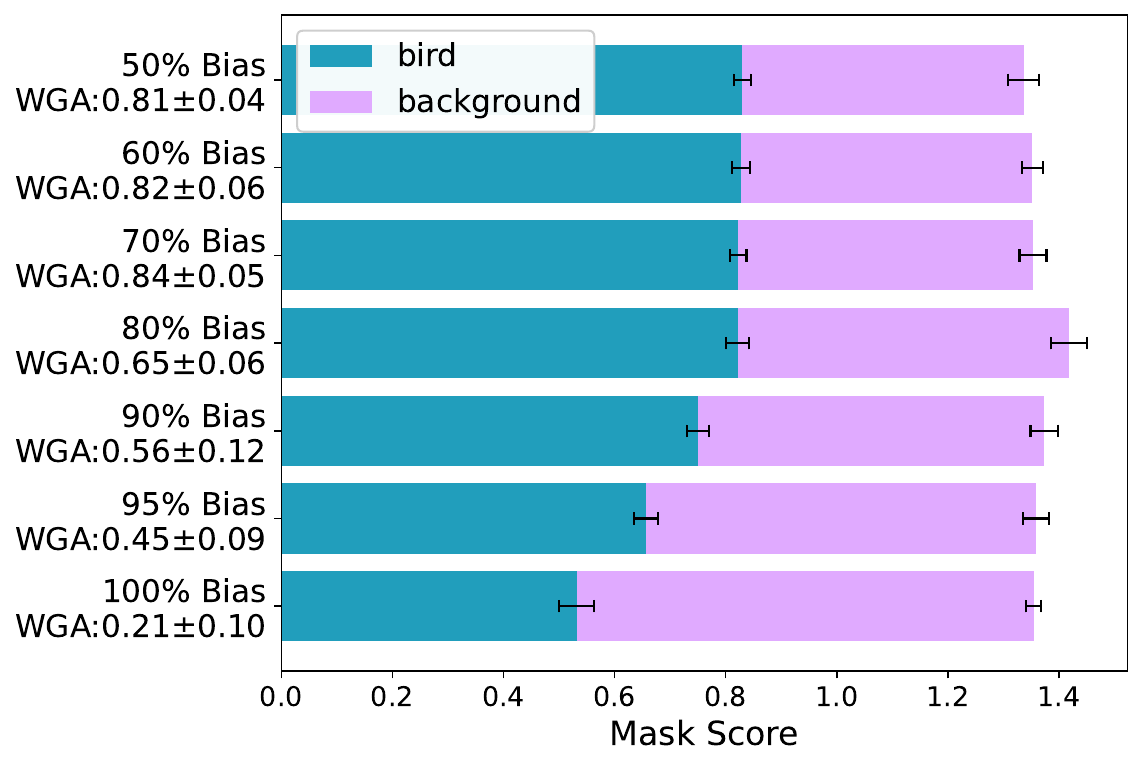}
    \centering
    \caption{\textbf{Evaluation of mask score using GradCAM on Waterbirds test set.} The X-axis represents the \metric{} mask score for the ground-truth masks of the bird and background. We note the dataset bias and the worst group accuracy (\textbf{WGA}) along the Y-axis. As the bias increases, the worst group accuracy decreases and the model attends less to the bird and more to the background.}
    \label{fig:waterbirds}
\end{figure}

\smallsec{Results} We compare the heatmap generated with the ground-truth masks for the bird. In \cref{fig:waterbirds_heatmaps}, we show the average bird mask, as well as the average heatmaps generated by GradCAM across all images in the test set for models trained at different levels of bias. As the bias increases, models rely more on cues from the background. This is reflected in the heatmaps, which highlight regions other than the bird mask.
We verify that \metric{} captures this effect in \cref{fig:waterbirds}, which shows the mask scores across varying training set bias for both bird and background masks. We also report the worst group accuracy (WGA) of models for each. As expected, the worst group accuracy decreases from $0.81\pm0.02$ to $0.21\pm0.10$ as bias increases from 50\% to 100\%. The decrease from $0.72\pm0.02$ to $0.42\pm0.03$ in mask score almost exactly mirrors the proportional decrease in WGA, validating that the metric accurately measures model bias. Due to the simple nature of Waterbirds, the bias in the dataset is directly represented in the training distribution, and \metric{} captures this perfectly.

\section{Analyzing CelebA}
In this section, we analyze the CelebA dataset~\cite{liu_deep_2015}
using \metric{}. CelebA is a widely used dataset for a variety of tasks, including evaluating debiasing methods. CelebA contains 2,022,599 images of celebrity faces, each labeled with 40 binary attributes, including both attributes localized to specific face regions (\eg, \texttt{Big\_Nose}, \texttt{Mouth\_Slightly\_Open}, \texttt{Blond\_Hair}) and attributes that are more global (\eg, \texttt{Male}\footnotemark, \texttt{Heavy\_Makeup}). We use \metric{} to understand more about the attributes in the dataset, and how they might influence each other.
\footnotetext{We acknowledge that these binary feature labels in CelebA, especially the \texttt{Male} label, forces people's presentations to fit into binaries. The \texttt{Male} label inherently assumes that an individual's gender presentation is tied to their gender identity. It is not clear what standards the creators of CelebA use in their definition of the \texttt{Male} label and other feature labels. However, for our goal of creating and evaluating bias metrics, we follow existing literature in our use of CelebA labels.}

\smallsec{Background}
The CelebA dataset is one of the most widely used benchmarks for studying facial recognition, debiasing, and generative modeling~\cite{liu_deep_2015}.  
Studies using CelebA have significantly advanced their respective fields. 
In generative modeling for example, CelebA is a common real world testbed, such as StarGAN for facial attribute transfer and in CoCosNetv2 for image translation~\cite{Choi_2018_CVPR, zhou_cocosnet_2021, xu_positional_2021}. 
The recent explosion of text-to-image models that can be personalized and controlled for realistic synthesis has caused a resurgence of facial recognition models for controllable editing~\cite{liu_stgan_2019}. 
Finally, many techniques for bias mitigation are validated on CelebA, from reweighting by using committees or biased models, to re-sampling or using pseudo-labels~\cite{kim_learning_2024, nam_learning_2020, xu_investigating_2020, seo_unsupervised_2022, qraitem_bias_2023}.
A commonly studied setting is \texttt{Blond\_Hair} as the target attribute and \texttt{Male} as the protected attribute, as popularized in the evaluation of group DRO paper by Sagawa \etal~\cite{sagawa_distributionally_2020} and used by many subsequent works \cite{kim_discovering_2024,jang_difficulty-based_2023,seo_unsupervised_2022}.

Several followups of the original dataset have also been developed for further study, such as the CelebA-HQ subset of 30,000 images of 1024×1024 resolution~\cite{karras_progressive_2018}, as well as the CelebAMask-HQ dataset which additionally annotates the images with semantic masks of 19 facial component categories at a $512\times512$ resolution~\cite{lee_maskgan_2020}.
The high resolution datasets are especially useful for testing high quality super-resolution and inpainting ~\cite{chen_learning_2021, zeng_learning_2019}.

Despite its popularity, CelebA has many flaws which have been noted in previous works. Several attributes (\eg, \texttt{Big\_Lips}, \texttt{Heavy\_Makeup}, etc.) have been shown to be inconsistently labelled~\cite{ramaswamy_fair_2021,qraitem_bias_2023}. Ramaswamy \etal also find that 13 of the attributes exhibit extreme class imbalance for gender expression~\cite{ramaswamy_fair_2021}. 
Others find issues of hidden (unlabeled) biases, which bias discovery works 
aim to target, such as hair length and visible hair area~\cite{balakrishnan_towards_2020, li_discover_2022}.
These issues in CelebA directly lead to biased models and generations. 
We aim to shed light on these different biases, to better understand how they occur and propagate into \mbox{trained models.} 

\subsection{Comparing to ground-truth masks}
\label{subsec:ground-truth}

We start by evaluating heatmaps using ground-truth masks, for attributes that are localized and have associated masks. 

\smallsec{Experiment Setup}
Since we require ground-truth segmentation masks, we use CelebAMask-HQ~\cite{lee_maskgan_2020}, a subset of CelebA in which each image has a high-quality segmentation mask of different facial features, including hair, nose, skin, hats, and jewelry. We group like features together, \eg, \{left brow, right brow\} and \{upper lip, lower lip, mouth\}. Large non-localized feature masks (background, skin, and cloth) are excluded from our analyses. We choose a 70\%-15\%-15\% train-validation-test split for training on CelebAMask-HQ.
To train classifiers for the attributes, we use a ResNet-50 model~\cite{he_deep_2016} pretrained on ImageNet~\cite{russakovsky_imagenet_2015}. We replaced the final layer with two fully-connected layers with a hidden layer size of 2,048 and a dropout layer between them in order to improve accuracy, following Ramaswamy \etal for their CelebA ResNet classifier~\cite{ramaswamy_fair_2021}. We used a binary cross-entropy loss, weighted proportionally to positive examples of each attribute, with a batch size of 32. Other hyperparameters remain the same as \cref{sec:validation}. 

\begin{figure}[t]
    \centering
    \includegraphics[width=\linewidth]{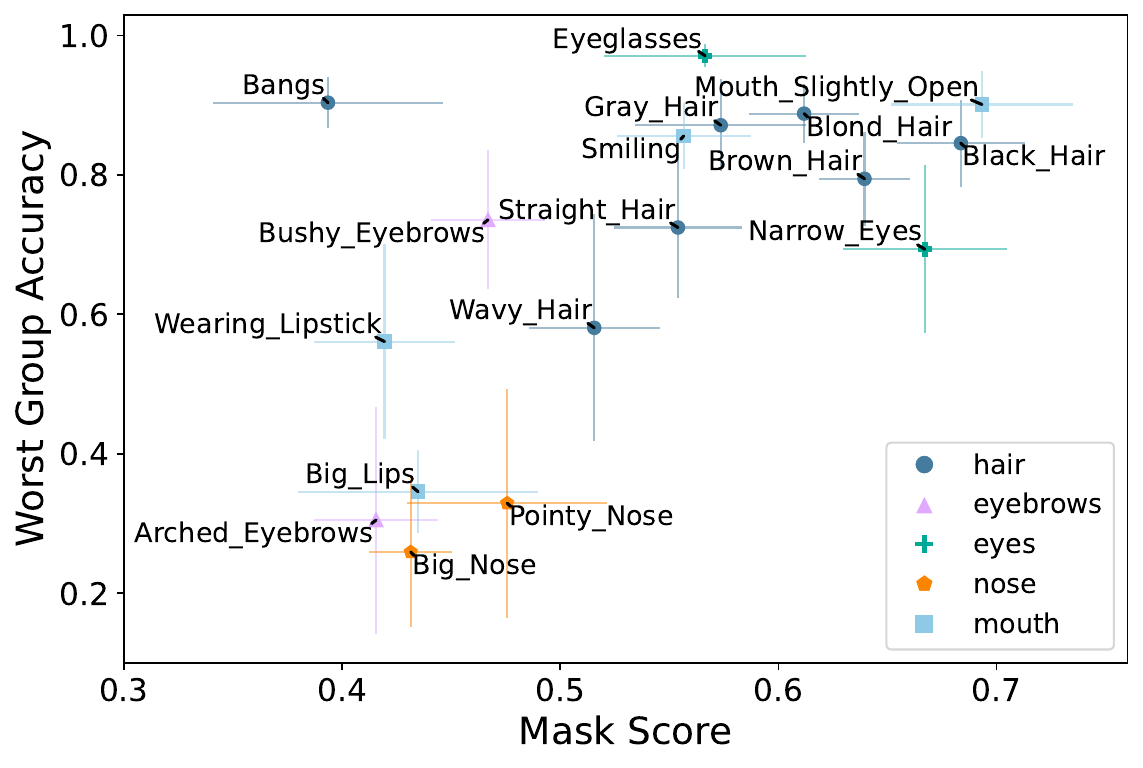}
    \caption{\textbf{Evaluation of mask score using GradCAM on CelebA test set with attribute-specific feature masks, compared to worst group accuracy with \texttt{Male}.} A mask score of 1 indicates perfect agreement between the attention map and feature mask, and 0 indicates perfect disagreement. Groups are considered based on ground-truth labels for the different combinations of target attribute and \texttt{Male}. If the number of images in a group is less than 1\% of the test set, the group was excluded from consideration.}
    \label{fig:mask_scatter}
\end{figure}

\smallsec{Results}
We choose a subset of 17 CelebA attributes that have directly corresponding feature masks, and calculate the respective mask score for each attribute (\cref{fig:mask_scatter}). Unlike Waterbirds, there is not a strong correlation between worst group accuracy (WGA) and the mask score. This is not surprising, since dataset bias is not immediately correlated to a singular attribute's labeling. Instead, an attribute's WGA and bias is dependent on the features in the image and the distribution of its label with the labels of other attributes. For example, \texttt{Wearing\_Lipstick} has a moderately high WGA, but a relatively low mask score. We hypothesize that this effect is due to the attribute's very strong correlation with \texttt{Male}, causing the model to attend away from the mouth and towards features relevant for \texttt{Male}. Other attributes, like \texttt{Eyeglasses}, have both a high mask score and WGA, because they are highly distinguishable. 
\nowidow

\subsection{Comparison with the \texttt{Male} heatmap}
\label{subsec:male-comparison}
In line with prior works, which investigate the impact of bias due to the protected \texttt{Male} attribute, we next examine the correlation between the heatmaps of different attributes and the heatmap for the \texttt{Male} attribute. The experimental setup remains the same as \cref{subsec:ground-truth}. 

\begin{figure*}[!t]
    \centering
    \begin{subfigure}{0.49\linewidth}
        \centering
        \includegraphics[width=\linewidth]{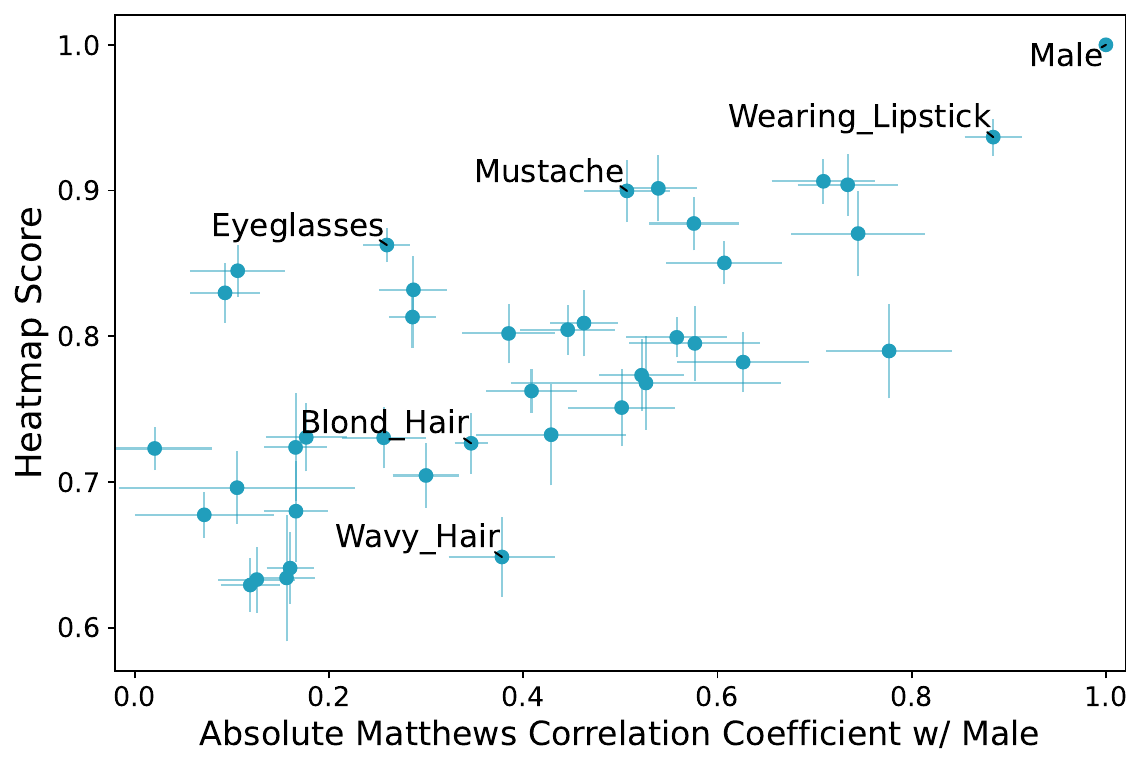}
    \end{subfigure}
    \begin{subfigure}{0.49\linewidth}
        \centering
        \includegraphics[width=\linewidth]{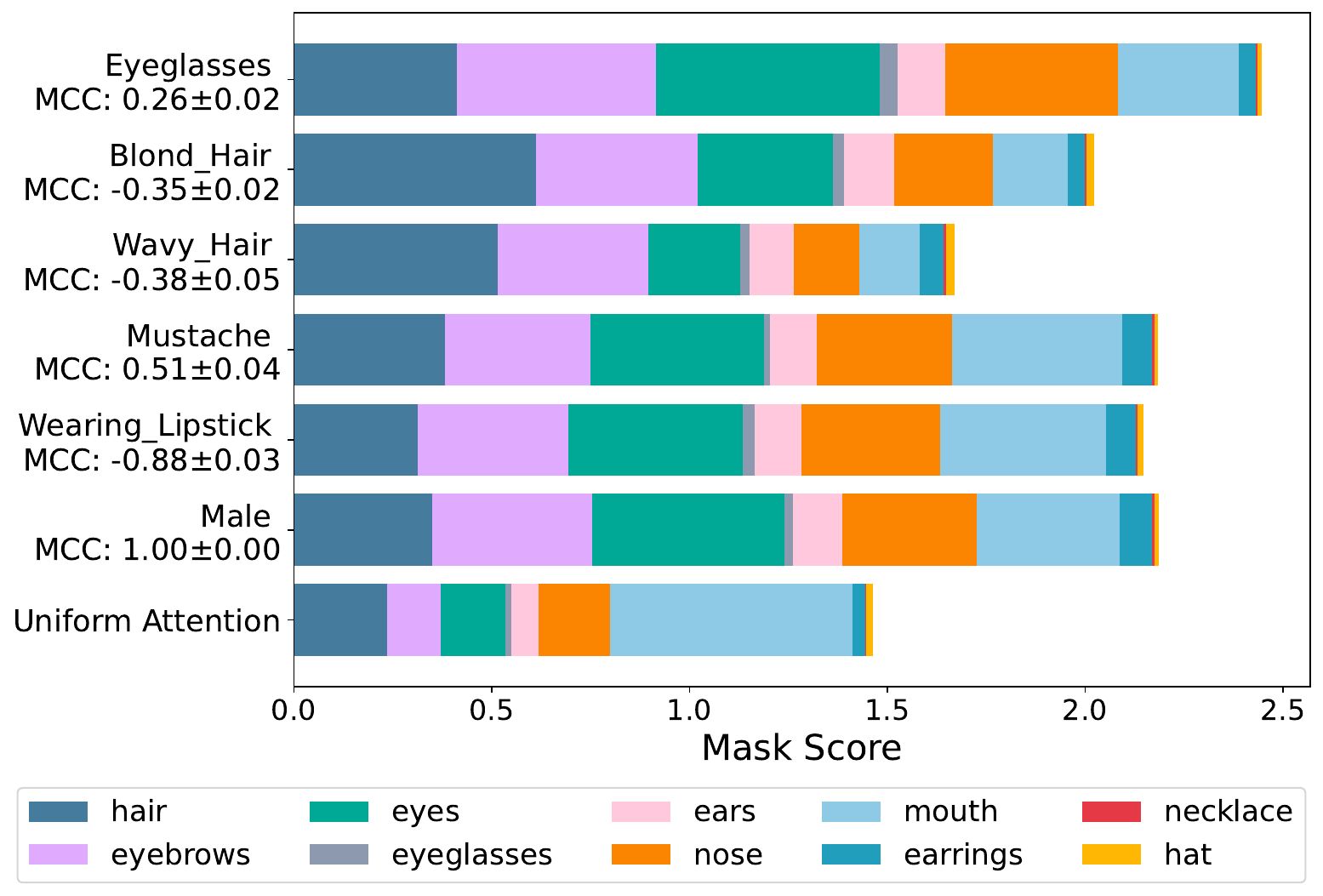}
    \end{subfigure}
    \caption{\textbf{Comparison of attributes with the \texttt{Male} attribute heatmap}. (\emph{Left}) We compare \metric{} with the absolute value of the Matthews correlation coefficient between the predictions of the attribute and \texttt{Male}, noticing a strong positive trend. Some attributes are outliers to this trend, including \texttt{Eyeglasses} and \texttt{Mustache}, which lie above this trend, and \texttt{Wavy\_Hair}, which lies below. (\emph{Right}) We measure the mask score for a selection of attributes. We notice that the heatmap for \texttt{Male} attends most strongly to the eye, eyebrows, and mouth region, which is closely mimicked by \texttt{Wearing\_Lipstick}. 
    We can also compare attributes like \texttt{Blond\_Hair} and \texttt{Wavy\_Hair}, and find that the main difference between their heatmaps is in the~eye~region.}
    \label{fig:celeb}
\end{figure*}

We compute \metric{} for all 40 attributes with \texttt{Male} (\cref{fig:celeb} \emph{left}). We measure the correlation between the attribute and the \texttt{Male} label using the absolute value of Matthews correlation coefficient (MCC), which is tailored for comparing two binary variables. The heatmap score ranges from $0.63\pm0.02$ for \texttt{Black\_Hair} to $0.94\pm0.01$ for \texttt{Wearing\_Lipstick}. \texttt{Male} is 1 because its attention map is being compared with itself. There is a clear positive trend between the heatmap score and predicted label MCC. Some attributes are outliers to this trend, such as \texttt{Mustache} and \texttt{Eyeglasses} having higher heatmap scores, and \texttt{Wavy\_Hair} having a lower heatmap score. We also report the mask score for selected attributes (\cref{fig:celeb} \emph{right}). The mask score for \texttt{Male} demonstrates that the models attend most strongly to the eye, eyebrow, and mouth region of the face, and slightly less to the nose and hair regions. We notice that this is most closely replicated by \texttt{Wearing\_Lipstick}, validating the high heatmap score. This per-region score computation also allows us to understand \textit{how} features of different attributes differ: for example, the main difference between \texttt{Blond\_Hair} and \texttt{Wavy\_Hair} appears to be in how much the models attend to regions around the eyes and nose. 


We now analyze in detail five attributes representative of those with distinct properties: 
\begin{itemize}
    \item \texttt{Wearing\_Lipstick}: This attribute is strongly correlated with \texttt{Male}, in both MCC and heatmap score.
    \item \texttt{Eyeglasses} and \texttt{Mustache}: These are outliers to the heatmap score trend, having significantly higher heatmap scores compared to other attributes with similar MCCs.
    \item  \texttt{Blond\_Hair} and \texttt{Wavy\_Hair}: This pair of attributes relate to the same regions within the image (hair) with similar MCCs, but have very different heatmap scores. 
\end{itemize}

\smallsec{Wearing Lipstick}
\texttt{Wearing\_Lipstick} has the highest absolute correlation with \texttt{Male} out of all 40 attributes, with an MCC of $0.88\pm0.03$. Furthermore, this correlation is predictive in both directions. One would expect that the attention map for \texttt{Wearing\_Lipstick} would highlight the mouth region. However, the mask score shows that the models attend to the eyes, eyebrows, nose, and hair regions, in addition to the mouth. In fact, the mask score distribution for \texttt{Wearing\_Lipstick} is closely similar to that of \texttt{Male}, only with a slightly higher mouth mask score. This close similarity between \texttt{Wearing\_Lipstick} and \texttt{Male} is reflected in the heatmap score, the highest of any attribute.
\nowidow

\smallsec{Eyeglasses}
    \texttt{Eyeglasses} is moderately correlated with \texttt{Male}, having an MCC of $0.26\pm0.02$, suggesting that \texttt{Male} is unlikely to influence the prediction of \texttt{Eyeglasses} much (or vice versa). As shown by the \texttt{Eyeglasses} mask score, the models attend strongly to the eyes, eyebrows, and nose regions. The score for the eyeglasses mask is low, because the score is averaged over all images in the test set, most of which do not contain eyeglasses as a mask. However, the eyeglasses mask score for \texttt{Eyeglasses} is still the highest for any attribute, suggesting that when \texttt{Eyeglasses} is present, the models attend highly to that region. Surprisingly for an attribute with a low MCC, the heatmap score for \texttt{Eyeglasses} is high at $0.86\pm0.01$. We posit that this might be due to one of the weaknesses within \metric{}: it's unable to detect when features are co-localized: we notice in \cref{fig:celeb} (\textit{right}) the heatmap attends highly to eyes and eyebrows, similar to that in \texttt{Male}. 
    \noclub

\begin{figure}[t]
    \centering
    \setlength{\tabcolsep}{2pt}
    \small{\begin{tabularx}{\linewidth} { 
          >{\centering\arraybackslash}X 
          >{\centering\arraybackslash}X 
          >{\centering\arraybackslash}X 
          >{\centering\arraybackslash}X}
    \includegraphics[width=\linewidth]{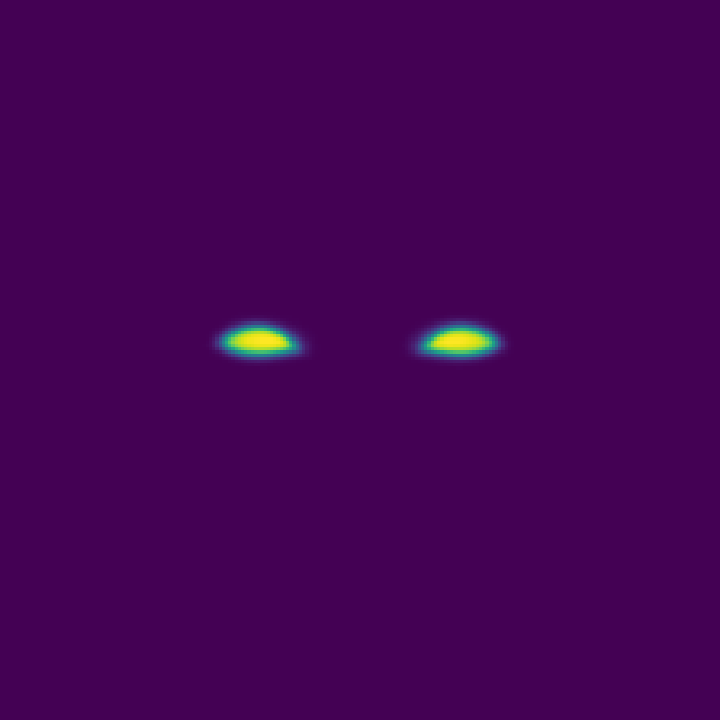} &
     \includegraphics[width=\linewidth]{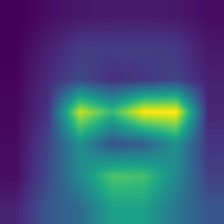} & 
         \includegraphics[width=\linewidth]{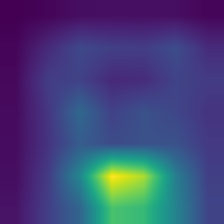}
         & \includegraphics[width=\linewidth]{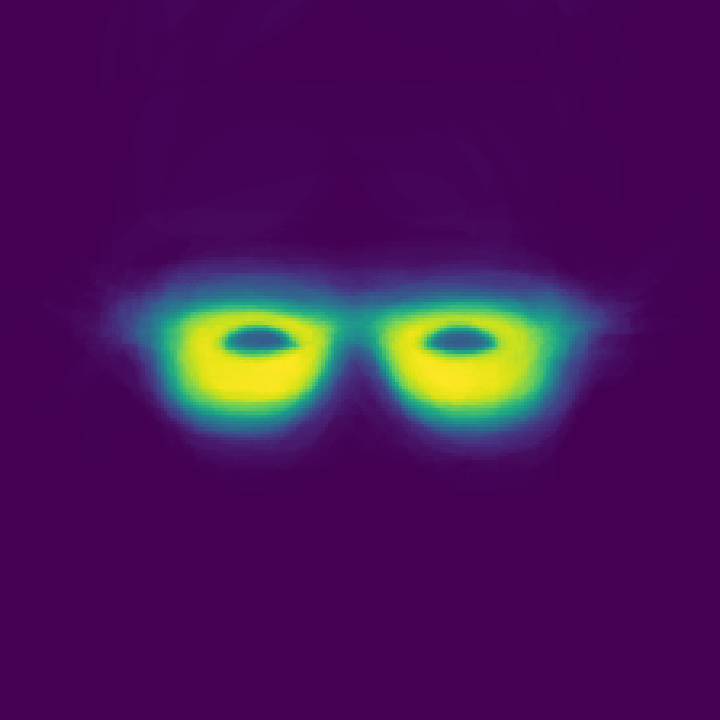} \\
    Eyes Mask& No \texttt{Eye}-\texttt{glasses} & \texttt{Eye}-\texttt{glasses} & Eyeglasses Mask \\
    \end{tabularx}}
    \caption{\textbf{Average heatmaps for \texttt{Male} with average masks.} We train models to predict \texttt{Male} when \texttt{Eyeglasses} are absent (\emph{center-left}) and present (\emph{center-right}). There is a stark difference in the heatmaps, suggesting that the features used by the model for predicting \texttt{Eyeglasses} are distinct from those used to predict \texttt{Male}, despite them being co-localized in the original models.}
    \label{fig:eyeglasses}
\end{figure}

    To verify this, we train two separate sets of models, one with just images for which \texttt{Eyeglasses} are present, and another for which \texttt{Eyeglasses} are absent. We hypothesize that if the \texttt{Male} and \texttt{Eyeglasses} classifiers are using the same features, \texttt{Male} would continue to attend to the eye region, since these features would continue to be useful. However, when \texttt{Eyeglasses} are present, \texttt{Male} attends primarily to the mouth, not the eyes, because eyeglasses are obscuring the features relevant to \texttt{Male} (\cref{fig:eyeglasses}). Thus, the high heatmap score \texttt{Eyeglasses} is not due to an underlying bias with \texttt{Male} in the model, but instead caused by co-localized features relevant to both attributes.
    \noclub

\smallsec{Mustache}
\texttt{Mustache} is moderately correlated with \texttt{Male}, with a predicted label MCC of $0.51\pm0.04$. \texttt{Mustache}'s mask score distribution reflects that of \texttt{Male}, with slightly more attention to the hair and mouth regions. This is reflected by a high heatmap score of $0.90\pm0.02$. We choose this attribute since this attribute represents a one-way correlation: images where \texttt{Mustache} are labeled as present are almost often labeled \texttt{Male}, whereas images where \texttt{Mustache} are labeled as absent are roughly evenly split among being labeled \texttt{Male} and not \texttt{Male}. 

\begin{figure}[t]
    \centering
    \setlength{\tabcolsep}{2pt}
    \small{\begin{tabularx}{\linewidth} { 
          >{\centering\arraybackslash}X  >{\centering\arraybackslash}X 
          >{\centering\arraybackslash}X  
          >{\centering\arraybackslash}X}
    \includegraphics[trim={0 0 0 20},clip,width=\linewidth]{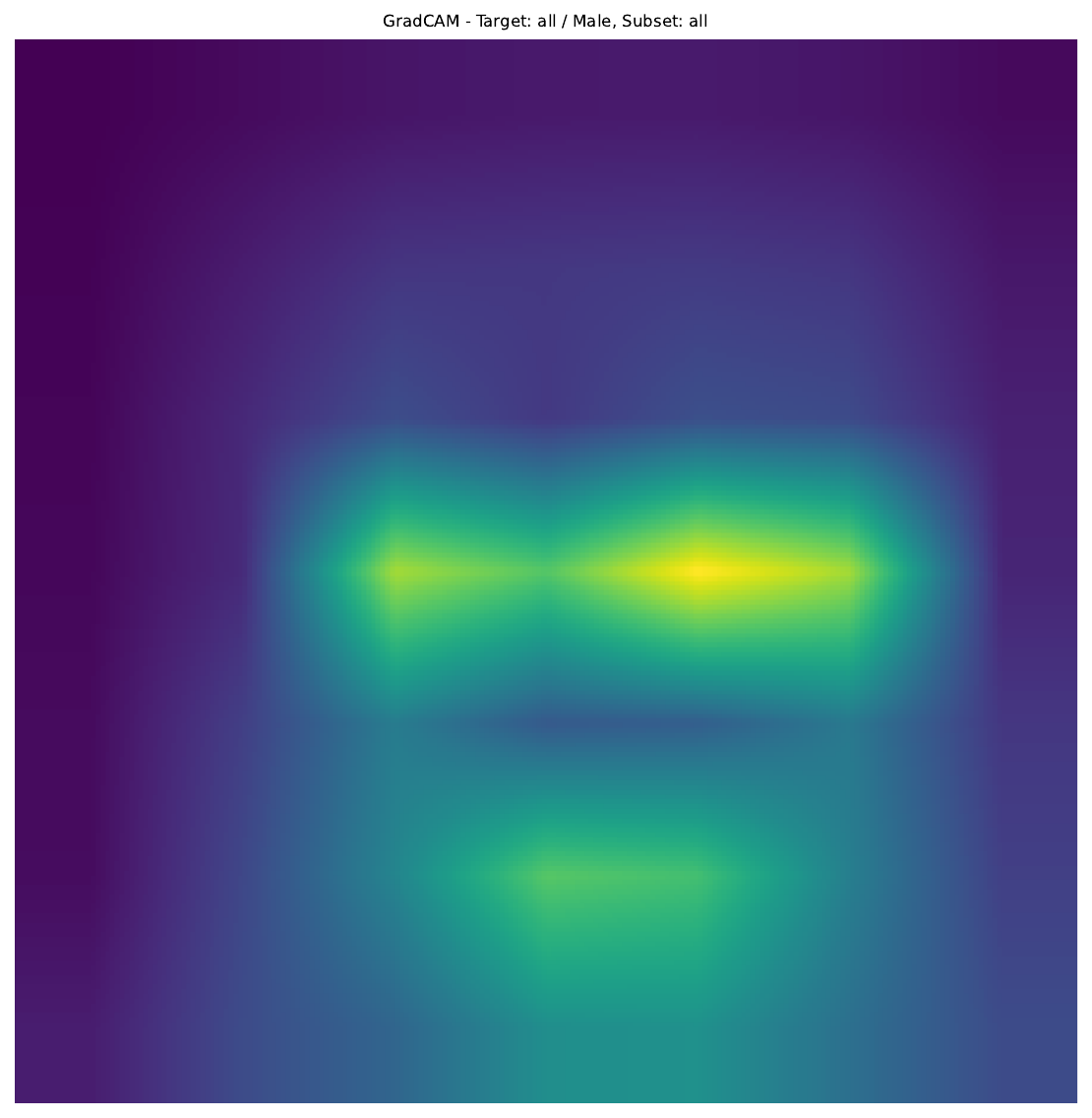} &
    \includegraphics[trim={0 0 0 20},clip,width=\linewidth]{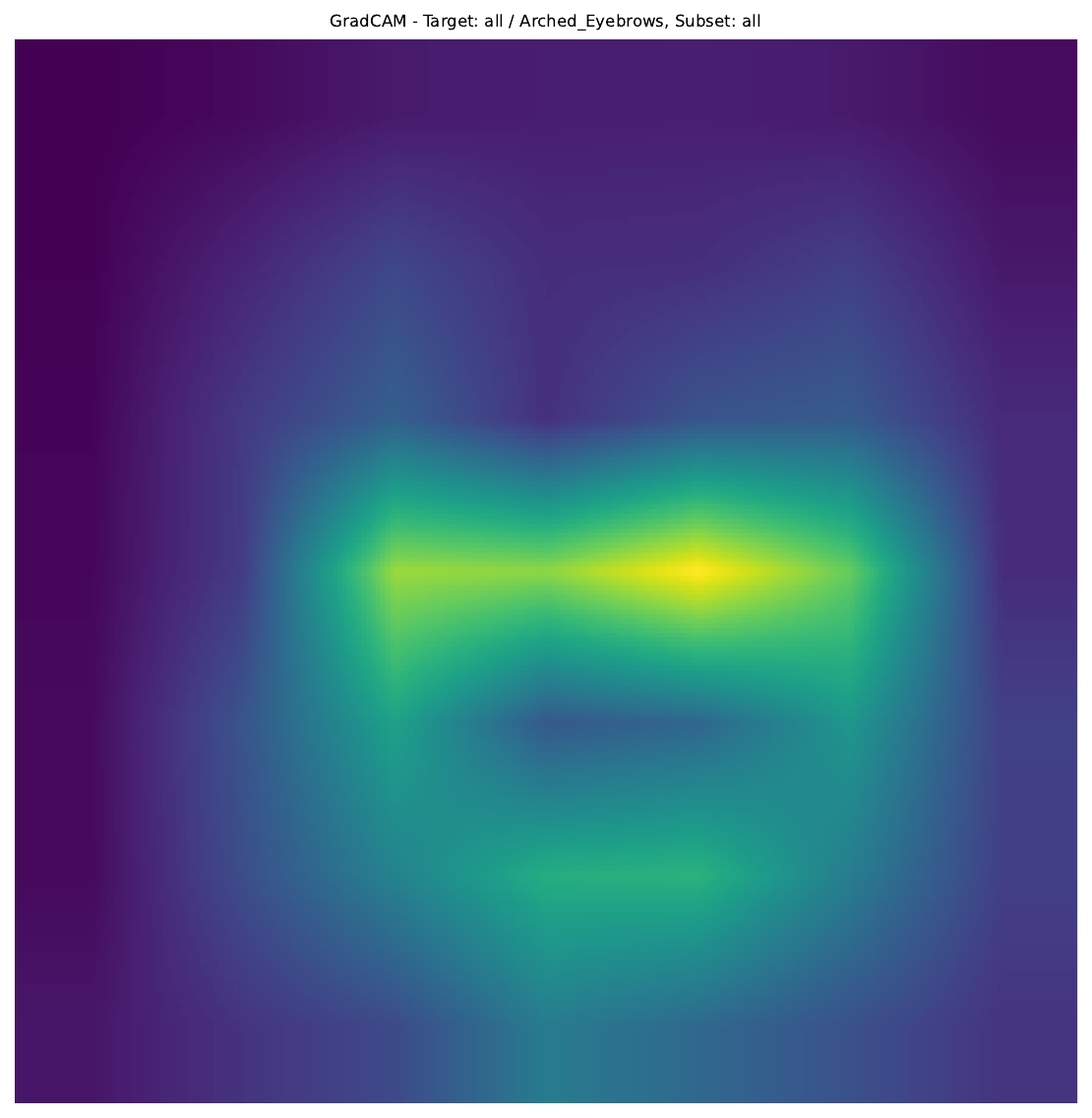} &
     \includegraphics[trim={0 0 0 20},clip,width=\linewidth]{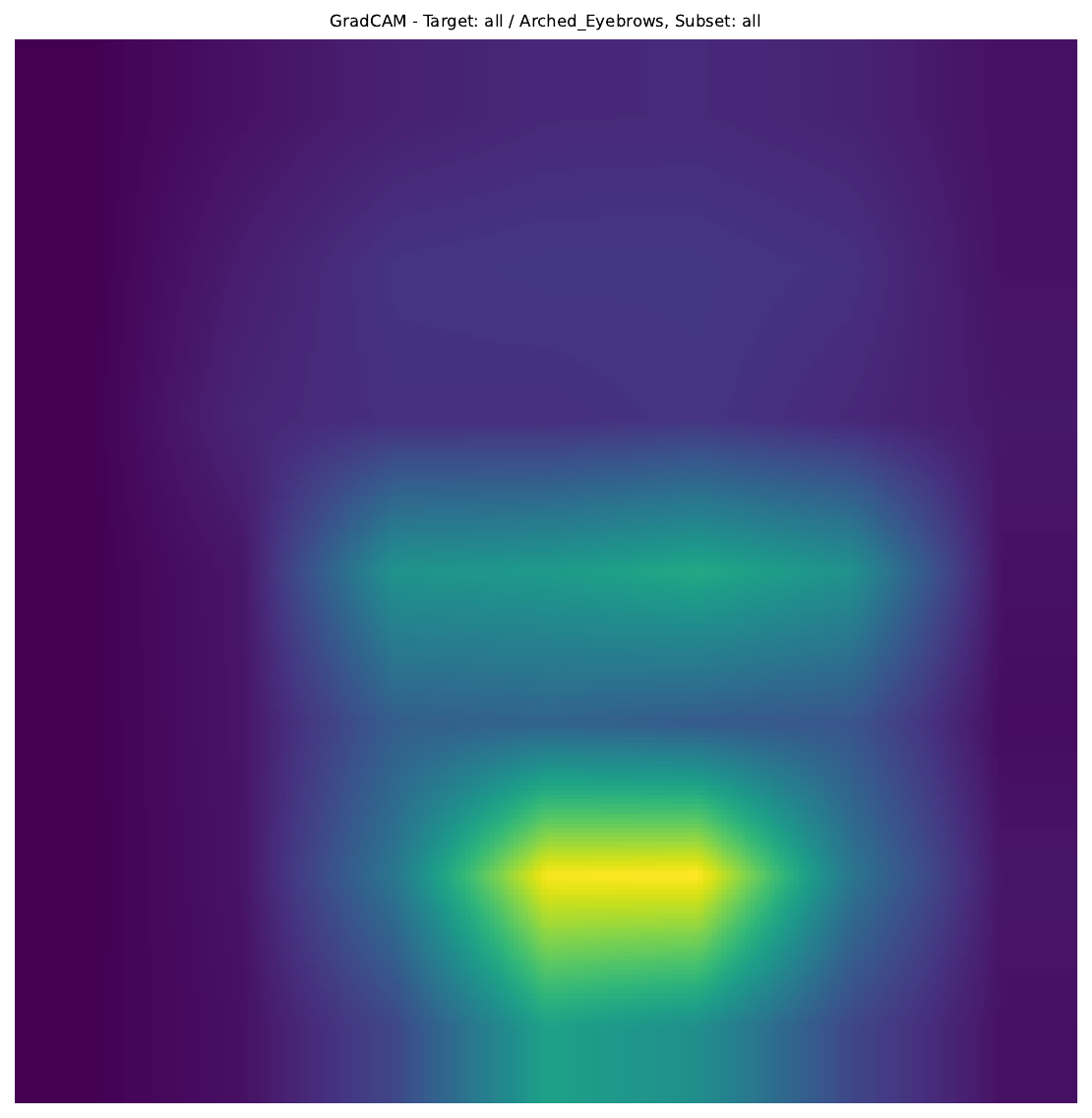} & 
         \includegraphics[trim={0 0 0 20},clip,width=\linewidth]{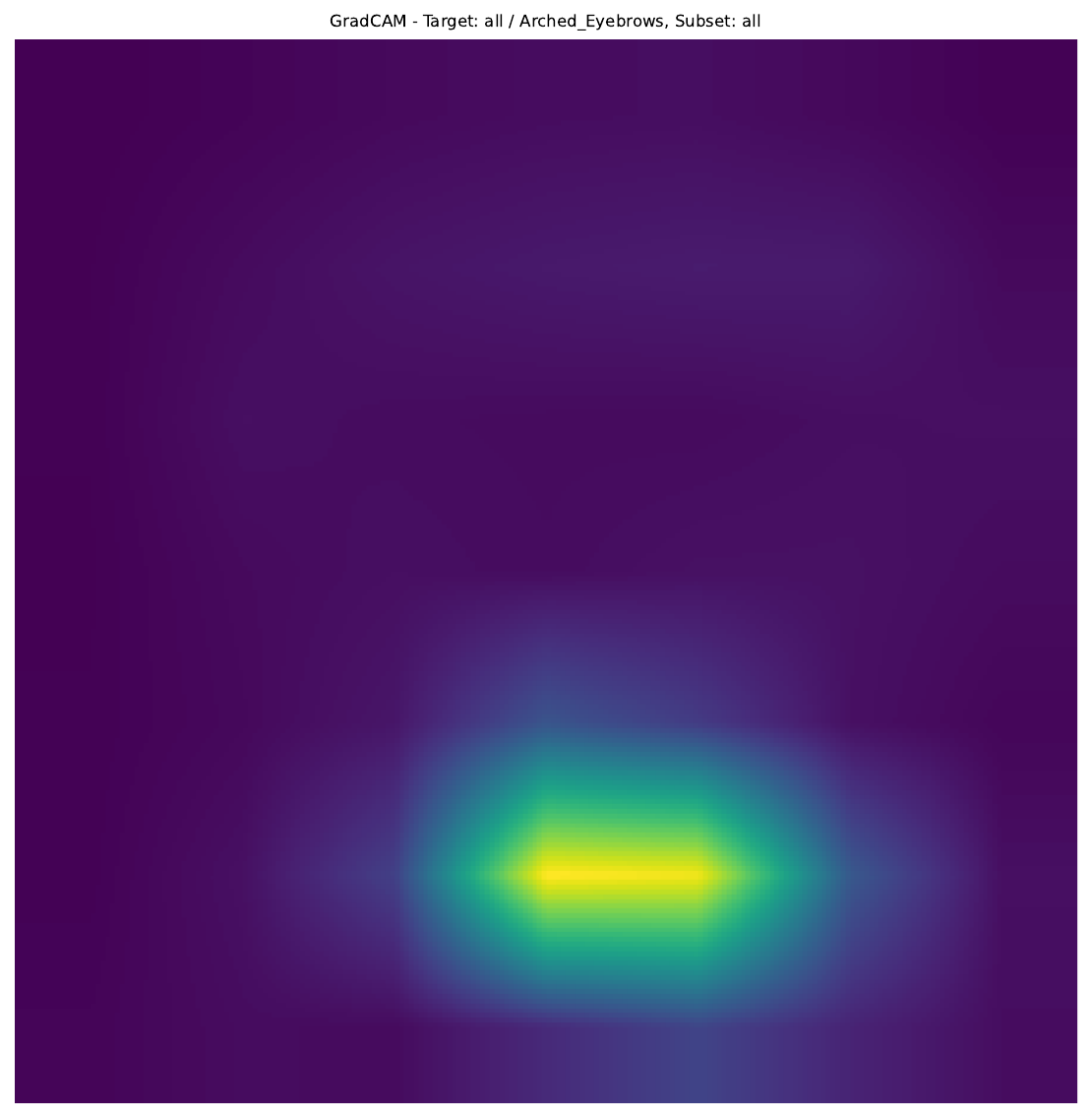} \\
     \texttt{Male} heatmap & not \texttt{Male}, no \texttt{Mustache} & \texttt{Male}, no \texttt{Mustache}  & \texttt{Male}, \texttt{Mustache} 
    \end{tabularx}}
    \caption{\textbf{Average heatmaps for \texttt{Mustache}}. We visualize average heatmaps for \texttt{Mustache} for images where \texttt{Mustache} and \texttt{Male} are labeled false (\textit{center-left}), where \texttt{Mustache} is labeled false and \texttt{Male} is labeled true (\textit{center-right}) and where \texttt{Mustache} and \texttt{Male} are labeled true (\emph{far right}), and compare to the \texttt{Male} heatmap (\emph{far left}). When \texttt{Male} is labeled as false, \texttt{Mustache} and \texttt{Male} attention maps closely align but do not when \texttt{Male} is labeled true.}
    \label{fig:mustache}
\end{figure}

We investigate how \metric{} changes based on the ground-truth values of these attributes (\cref{fig:mustache}). The score is extremely high ($0.94\pm0.0$2) among images labeled not \texttt{Male}. When \texttt{Male} is false, the \texttt{Mustache} and \texttt{Male} attention maps closely align, indicating that the model is heavily relying on \texttt{Male} to classify \texttt{Mustache}. However, when the image is labeled as \texttt{Male}, the score is lower ($0.84\pm0.5$ and $0.82\pm0.03$ for \texttt{Mustache} true and false respectively), the models attend less to \texttt{Male} regions in order to classify \texttt{Mustache}. \texttt{Mustache} demonstrates that even though two attributes may be one-way predictive in the dataset (and thus have a lower MCC), the models still attend strongly to any correlation between the attributes, which is indicated through \metric{}. 
    

\smallsec{Blond Hair and Wavy Hair}
Both \texttt{Blond\_Hair} and \texttt{Wavy\_Hair} have similar predicted label MCCs of $0.34\pm0.02$ and $0.37\pm0.05$ respectively. Despite both referring to the hair feature, \texttt{Blond\_Hair} and \texttt{Wavy\_Hair} exhibit distinct attention maps. Relative to the \texttt{Male} mask score, for \texttt{Wavy\_Hair} the models attend to more to the hair region, and significantly less to the eyes, nose, and mouth. This increase for hair is larger regarding \texttt{Blond\_Hair}, which also has a smaller decrease in the eye region. Overall, \texttt{Blond\_Hair} has a higher heatmap score of $0.72\pm0.02$, while \texttt{Wavy\_Hair} is lower at $0.65\pm0.03$. 

We investigate difference further, positing two potential hypotheses: first, the \texttt{Wavy\_Hair} has a significantly lower \apnorm{} of $ 0.80\pm0.03$, compared to $0.96\pm0.01$ for \texttt{Blond\_Hair}. This could be due to labeling inconsistencies for \texttt{Wavy\_Hair}~\cite{ramaswamy_fair_2021, bebis_quantitative_2022} resulting in the heatmaps being less useful for this attribute, since GradCAM uses the predicted label to generate heatmaps. Another hypothesis for this difference could be that one of these attributes are not directly related with the \texttt{Male} attribute, instead, the attribute and \texttt{Male} are both correlated with an (unlabeled) confounder attribute, resulting in this correlation. 

 \begin{figure}[t]
        \centering
        \begin{subfigure}{0.515\linewidth}
            \centering
            \caption*{\quad\ \texttt{Blond\_Hair} $\tau$: 0.073}
            \includegraphics[width=\linewidth]{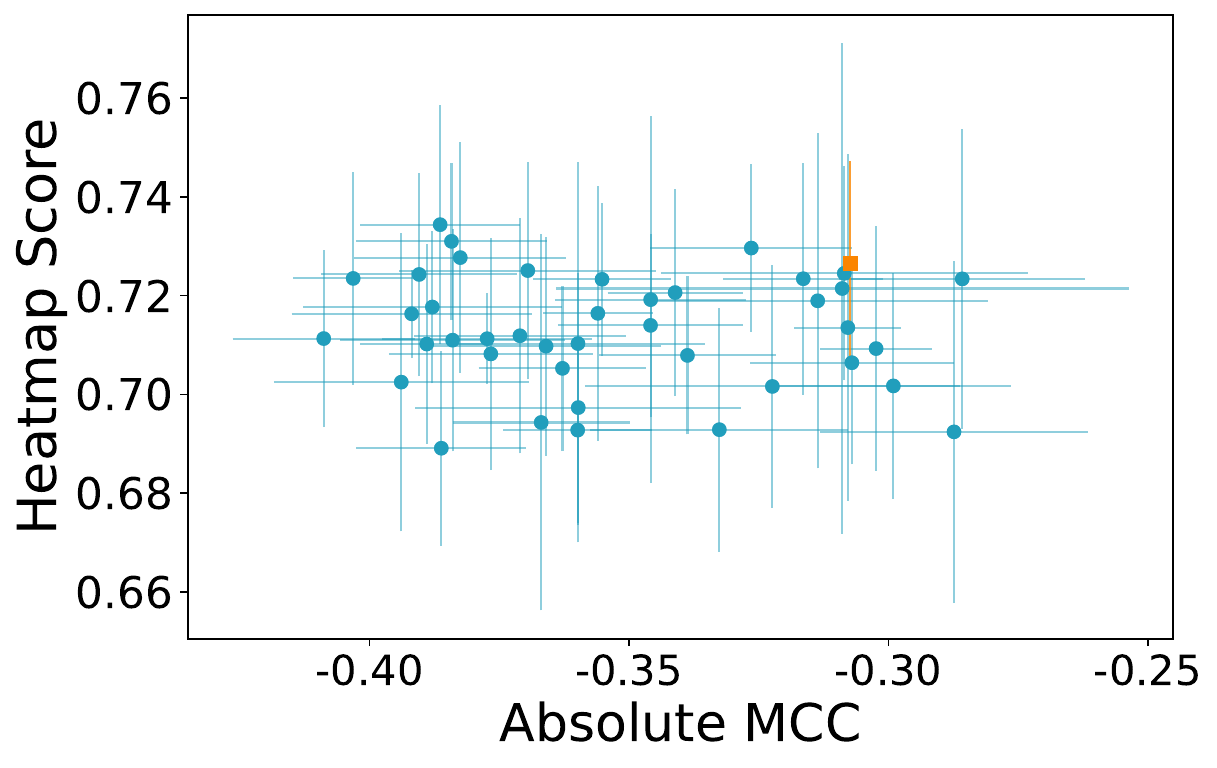}
        \end{subfigure}\!
        \begin{subfigure}{0.481\linewidth}
            \centering
            \caption*{\quad\ \texttt{Wavy\_Hair} $\tau$ 0.778}
            \includegraphics[width=\linewidth]{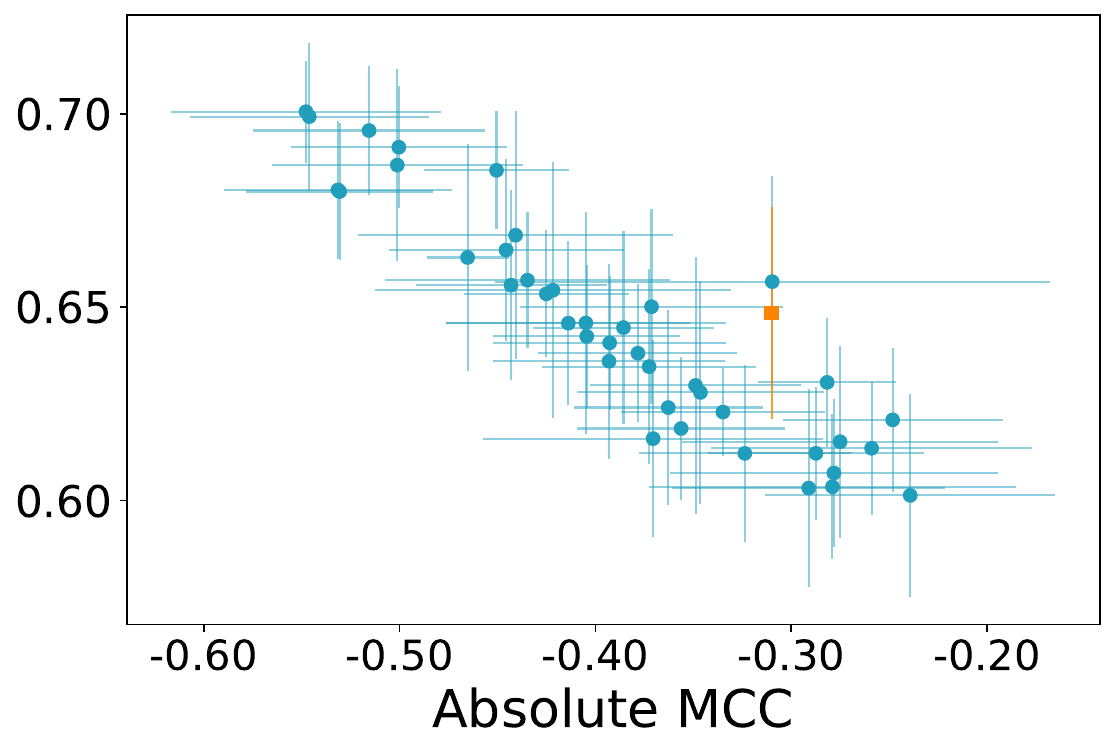}
        \end{subfigure}
        \caption{\textbf{Varying the correlation in the training dataset}. To understand if the correlations are indeed responsible for the mask scores in their entirety, we subsample the dataset to vary the ground-truth MCC between \texttt{Blond\_Hair} and \texttt{Wavy\_Hair} and \texttt{Male}. We find that changing the ground-truth MCC for \texttt{Blond\_Hair} (\emph{left}) does not change the heatmap score, while changing the MCC for \texttt{Wavy\_Hair} (\emph{right}) results in a strong change in the heatmap score (orange/square indicates the original results). This suggests that there might be a hidden confounder present between \texttt{Blond\_Hair} and \texttt{Male}, which leads to the large heatmap score. This is unlike \texttt{Wavy\_Hair}, which is much more dependent on ground-truth correlations within the dataset.}
        \label{fig:subgroup}
    \end{figure}
    
 To test this hypothesis, we modified the training distribution for \texttt{Blond\_Hair} and \texttt{Wavy\_Hair} by training models on a subsampled training set (\cref{fig:subgroup}). To do so, we varied the ground-truth MCC from $-0.5$ to $-0.1$ between the target attribute and \texttt{Male} by varying proportion of the 4 subgroups within the training set, keeping the overall number constant (details in \cref{sec:subgroup-details}). For \texttt{Blond\_Hair} we find that there is no statistically significant change in heatmap score, with a Kendall $\tau$ value of $0.007$. However, \texttt{Wavy\_Hair} demonstrates a strong correlation between MCC and heatmap score ($\tau = 0.785$), with the model bias decreasing as train set bias decreases. This indicates that there might be an unlabeled confounder present in \texttt{Blond\_Hair}: there is an innate quality to the features distinct from dataset labels that create bias within the model for \texttt{Blond\_Hair}, rather than the simple proportion of attributes to one another in the dataset as in \texttt{Wavy\_Hair}.

\section{Conclusion}
\label{sec:conclusion}

\metric{} yields several insights into the CelebA dataset~\cite{liu_deep_2015}. In particular, we identify specific ways in which different attributes are influenced by the \texttt{Male} label: attributes can be biased more or less based on labels of the sensitive attribute and can be biased in ways beyond the correlation of labels within the dataset. These insights allow us to better understand how debiasing techniques might perform on this dataset. For example, methods that attempt to rebalance the dataset or improve group accuracies for \texttt{Blond\_Hair}~\cite{sagawa_distributionally_2020, seo_unsupervised_2022} might struggle since the bias is not due to the presence of blond hair, but a hidden confounder. 

In conclusion, we propose \metric{}, a metric for identifying and explaining spurious correlations through attention maps. We demonstrate the metric's effectiveness through validations on the Waterbirds~\cite{sagawa_distributionally_2020} and CelebA~\cite{liu_deep_2015} datasets. Within CelebA, we show that the metric and the mask and heatmap scores reveals aspects beyond dataset labels and model accuracies, recontextualizing prior analyses of CelebA. Future investigations of the proposed methods on other datasets and tasks can provide further insights into the nature of biases within computer vision models.

\section*{Acknowledgements}
\label{sec:acknowledgements}

We acknowledge support from the Princeton SEAS Innovation Grant to VVR, and from the Princeton University's Office of Undergraduate Research Undergraduate Fund for Academic Conferences through the Hewlett Foundation Fund to AS. This material is based upon work supported by the National Science Foundation under grant No. 2145198 to OR. Any opinions, findings, and conclusions or recommendations expressed in this material are those of the authors and do not necessarily reflect the views of the National Science Foundation. The experiments presented in this work were performed on computational resources managed and supported by Princeton Research Computing, a consortium of groups including the Princeton Institute for Computational Science and Engineering (\mbox{PICSciE}) and Research Computing at Princeton University. \\
{
    \small
    \bibliographystyle{ieeenat_fullname}
    \bibliography{main}

\begin{thebibliography}{94}
\providecommand{\natexlab}[1]{#1}
\providecommand{\url}[1]{\texttt{#1}}
\expandafter\ifx\csname urlstyle\endcsname\relax
  \providecommand{\doi}[1]{doi: #1}\else
  \providecommand{\doi}{doi: \begingroup \urlstyle{rm}\Url}\fi

\bibitem[Asgari et~al.(2022)Asgari, Khani, Khani, Gholami, Tran, Mahdavi~Amiri, and Hamarneh]{asgari_masktune_2022}
Saeid Asgari, Aliasghar Khani, Fereshte Khani, Ali Gholami, Linh Tran, Ali Mahdavi~Amiri, and Ghassan Hamarneh.
\newblock {MaskTune}: {Mitigating} {Spurious} {Correlations} by {Forcing} to {Explore}.
\newblock In \emph{Advances in {Neural} {Information} {Processing} {Systems}}, pages 23284--23296. 2022.

\bibitem[Balakrishnan et~al.(2021)Balakrishnan, Xiong, Xia, and Perona]{balakrishnan_towards_2020}
Guha Balakrishnan, Yuanjun Xiong, Wei Xia, and Pietro Perona.
\newblock Towards {Causal} {Benchmarking} of {Bias} in {Face} {Analysis} {Algorithms}.
\newblock In \emph{Deep {Learning}-{Based} {Face} {Analytics}}, pages 327--359. Springer International Publishing, Cham, 2021.

\bibitem[Bang et~al.(2024)Bang, Boggust, and Satyanarayan]{bang_explanation_2024}
Hyemin Bang, Angie Boggust, and Arvind Satyanarayan.
\newblock Explanation {Alignment}: {Quantifying} the {Correctness} of {Model} {Reasoning} {At} {Scale}.
\newblock In \emph{{ECCV} {Workshop} on {Explainable} {Computer} {Vision}}, 2024.

\bibitem[Bellamy et~al.(2019)Bellamy, Dey, Hind, Hoffman, Houde, Kannan, Lohia, Martino, Mehta, Mojsilović, Nagar, Ramamurthy, Richards, Saha, Sattigeri, Singh, Varshney, and Zhang]{bellamy_ai_2019}
R.~K.~E. Bellamy, K. Dey, M. Hind, S.~C. Hoffman, S. Houde, K. Kannan, P. Lohia, J. Martino, S. Mehta, A. Mojsilović, S. Nagar, K.~Natesan Ramamurthy, J. Richards, D. Saha, P. Sattigeri, M. Singh, K.~R. Varshney, and Y. Zhang.
\newblock {AI} {Fairness} 360: {An} extensible toolkit for detecting and mitigating algorithmic bias.
\newblock \emph{IBM Journal of Research and Development}, 63:\penalty0 4:1--4:15, 2019.
\newblock Conference Name: IBM Journal of Research and Development.

\bibitem[Bianchi et~al.(2023)Bianchi, Kalluri, Durmus, Ladhak, Cheng, Nozza, Hashimoto, Jurafsky, Zou, and Caliskan]{bianchi_easily_2023}
Federico Bianchi, Pratyusha Kalluri, Esin Durmus, Faisal Ladhak, Myra Cheng, Debora Nozza, Tatsunori Hashimoto, Dan Jurafsky, James Zou, and Aylin Caliskan.
\newblock Easily {Accessible} {Text}-to-{Image} {Generation} {Amplifies} {Demographic} {Stereotypes} at {Large} {Scale}.
\newblock In \emph{2023 {ACM} {Conference} on {Fairness}, {Accountability}, and {Transparency}}, pages 1493--1504, 2023.

\bibitem[Buolamwini and Gebru(2018)]{buolamwini_gender_2018}
Joy Buolamwini and Timnit Gebru.
\newblock Gender {Shades}: {Intersectional} {Accuracy} {Disparities} in {Commercial} {Gender} {Classification}.
\newblock In \emph{Proceedings of the 1st {Conference} on {Fairness}, {Accountability} and {Transparency}}, pages 77--91. 2018.

\bibitem[Caliskan et~al.(2017)Caliskan, Bryson, and Narayanan]{caliskan_semantics_2017}
Aylin Caliskan, Joanna~J. Bryson, and Arvind Narayanan.
\newblock Semantics derived automatically from language corpora contain human-like biases.
\newblock \emph{Science}, 356\penalty0 (6334):\penalty0 183--186, 2017.

\bibitem[Caton and Haas(2024)]{caton_fairness_2024}
Simon Caton and Christian Haas.
\newblock Fairness in {Machine} {Learning}: {A} {Survey}.
\newblock \emph{ACM Computing Surveys}, 56\penalty0 (7):\penalty0 1--38, 2024.

\bibitem[Chen et~al.(2021)Chen, Liu, and Wang]{chen_learning_2021}
Yinbo Chen, Sifei Liu, and Xiaolong Wang.
\newblock Learning {Continuous} {Image} {Representation} with {Local} {Implicit} {Image} {Function}.
\newblock \emph{2021 IEEE/CVF Conference on Computer Vision and Pattern Recognition (CVPR)}, pages 8624--8634, 2021.

\bibitem[Choi et~al.(2018)Choi, Choi, Kim, Ha, Kim, and Choo]{Choi_2018_CVPR}
Yunjey Choi, Minje Choi, Munyoung Kim, Jung-Woo Ha, Sunghun Kim, and Jaegul Choo.
\newblock Stargan: Unified generative adversarial networks for multi-domain image-to-image translation.
\newblock In \emph{Proceedings of the IEEE Conference on Computer Vision and Pattern Recognition (CVPR)}, 2018.

\bibitem[Chouldechova(2017)]{chouldechova_fair_2017}
Alexandra Chouldechova.
\newblock Fair {Prediction} with {Disparate} {Impact}: {A} {Study} of {Bias} in {Recidivism} {Prediction} {Instruments}.
\newblock \emph{Big Data}, 5\penalty0 (2):\penalty0 153--163, 2017.

\bibitem[de~Vries et~al.(2019)de~Vries, Misra, Wang, and van~der Maaten]{devries2019objectrecognition}
Terrance de Vries, Ishan Misra, Changhan Wang, and Laurens van~der Maaten.
\newblock Does {Object} {Recognition} {Work} for {Everyone}?
\newblock In \emph{Proceedings of the {IEEE}/{CVF} {Conference} on {Computer} {Vision} and {Pattern} {Recognition} ({CVPR}) {Workshops}}, 2019.

\bibitem[Denton et~al.(2020)Denton, Hutchinson, Mitchell, Gebru, and Zaldivar]{denton_image_2020}
Remi Denton, Ben Hutchinson, Margaret Mitchell, Timnit Gebru, and Andrew Zaldivar.
\newblock Image {Counterfactual} {Sensitivity} {Analysis} for {Detecting} {Unintended} {Bias}, 2020.
\newblock arXiv:1906.06439.

\bibitem[Dixon et~al.(2018)Dixon, Li, Sorensen, Thain, and Vasserman]{dixon_measuring_2018}
Lucas Dixon, John Li, Jeffrey Sorensen, Nithum Thain, and Lucy Vasserman.
\newblock Measuring and {Mitigating} {Unintended} {Bias} in {Text} {Classification}.
\newblock In \emph{Proceedings of the 2018 {AAAI}/{ACM} {Conference} on {AI}, {Ethics}, and {Society}}, pages 67--73, 2018.

\bibitem[Fabbrizzi et~al.(2022)Fabbrizzi, Papadopoulos, Ntoutsi, and Kompatsiaris]{fabbrizzi_survey_2022}
Simone Fabbrizzi, Symeon Papadopoulos, Eirini Ntoutsi, and Ioannis Kompatsiaris.
\newblock A survey on bias in visual datasets.
\newblock \emph{Computer Vision and Image Understanding}, 223:\penalty0 103552, 2022.

\bibitem[Fong et~al.(2019)Fong, Patrick, and Vedaldi]{fong2019extremal}
Ruth Fong, Mandela Patrick, and Andrea Vedaldi.
\newblock Understanding deep networks via extremal perturbations and smooth masks.
\newblock In \emph{ICCV}, 2019.

\bibitem[Gavrikov and Keuper(2024)]{gavrikov_can_2024}
Paul Gavrikov and Janis Keuper.
\newblock Can {Biases} in {ImageNet} {Models} {Explain} {Generalization}?
\newblock In \emph{2024 {IEEE}/{CVF} {Conference} on {Computer} {Vision} and {Pattern} {Recognition} ({CVPR})}, pages 22184--22194, 2024.

\bibitem[Geirhos et~al.(2019)Geirhos, Rubisch, Michaelis, Bethge, Wichmann, and Brendel]{geirhos_imagenet-trained_2019}
Robert Geirhos, Patricia Rubisch, Claudio Michaelis, Matthias Bethge, Felix~A. Wichmann, and Wieland Brendel.
\newblock {ImageNet}-trained {CNNs} are biased towards texture; increasing shape bias improves accuracy and robustness.
\newblock In \emph{International {Conference} on {Learning} {Representations}}, 2019.

\bibitem[Goyal et~al.(2019)Goyal, Wu, Ernst, Batra, Parikh, and Lee]{goyal2019counterfactual}
Yash Goyal, Ziyan Wu, Jan Ernst, Dhruv Batra, Devi Parikh, and Stefan Lee.
\newblock Counterfactual visual explanations.
\newblock In \emph{ICML}, 2019.

\bibitem[Hardt et~al.(2016)Hardt, Price, Price, and Srebro]{hardt_equality_2016}
Moritz Hardt, Eric Price, Eric Price, and Nati Srebro.
\newblock Equality of {Opportunity} in {Supervised} {Learning}.
\newblock In \emph{Advances in {Neural} {Information} {Processing} {Systems}}. 2016.

\bibitem[Harrison et~al.(2023)Harrison, Gualdoni, and Boleda]{harrison_run_2023}
Sophia Harrison, Eleonora Gualdoni, and Gemma Boleda.
\newblock Run {Like} a {Girl}! {Sport}-{Related} {Gender} {Bias} in {Language} and {Vision}.
\newblock In \emph{Findings of the {Association} for {Computational} {Linguistics}: {ACL} 2023}, pages 14093--14103, 2023.

\bibitem[He et~al.(2016)He, Zhang, Ren, and Sun]{he_deep_2016}
Kaiming He, Xiangyu Zhang, Shaoqing Ren, and Jian Sun.
\newblock Deep {Residual} {Learning} for {Image} {Recognition}.
\newblock In \emph{Proceedings of the {IEEE} {Conference} on {Computer} {Vision} and {Pattern} {Recognition} ({CVPR})}, 2016.

\bibitem[Hendricks et~al.(2018)Hendricks, Burns, Saenko, Darrell, and Rohrbach]{burns_women_2019}
Lisa~Anne Hendricks, Kaylee Burns, Kate Saenko, Trevor Darrell, and Anna Rohrbach.
\newblock Women {Also} {Snowboard}: {Overcoming} {Bias} in {Captioning} {Models}.
\newblock In \emph{Computer {Vision} – {ECCV} 2018}, pages 793--811. Springer International Publishing, Cham, 2018.

\bibitem[Hirota et~al.(2022)Hirota, Nakashima, and Garcia]{hirota_gender_2022}
Yusuke Hirota, Yuta Nakashima, and Noa Garcia.
\newblock Gender and {Racial} {Bias} in {Visual} {Question} {Answering} {Datasets}.
\newblock In \emph{2022 {ACM} {Conference} on {Fairness}, {Accountability}, and {Transparency}}, pages 1280--1292, 2022.

\bibitem[Hoiem et~al.(2012)Hoiem, Chodpathumwan, and Dai]{hoiem_diagnosing_2012}
Derek Hoiem, Yodsawalai Chodpathumwan, and Qieyun Dai.
\newblock Diagnosing {Error} in {Object} {Detectors}.
\newblock In \emph{Computer {Vision} – {ECCV} 2012}, pages 340--353, 2012.

\bibitem[Huang et~al.(2023)Huang, Wang, Liang, Deng, Shi, Wen, Zhang, and Zhao]{huang_gradient_2023}
Linzhi Huang, Mei Wang, Jiahao Liang, Weihong Deng, Hongzhi Shi, Dongchao Wen, Yingjie Zhang, and Jian Zhao.
\newblock Gradient {Attention} {Balance} {Network}: {Mitigating} {Face} {Recognition} {Racial} {Bias} via {Gradient} {Attention}.
\newblock In \emph{2023 {IEEE}/{CVF} {Conference} on {Computer} {Vision} and {Pattern} {Recognition} {Workshops} ({CVPRW})}, pages 38--47, 2023.

\bibitem[Izmailov et~al.(2024)Izmailov, Kirichenko, Gruver, and Wilson]{izmailov_feature_2024}
Pavel Izmailov, Polina Kirichenko, Nate Gruver, and Andrew~Gordon Wilson.
\newblock On feature learning in the presence of spurious correlations.
\newblock In \emph{Proceedings of the 36th {International} {Conference} on {Neural} {Information} {Processing} {Systems}}, pages 38516--38532, 2024.

\bibitem[Jang and Wang(2023)]{jang_difficulty-based_2023}
Taeuk Jang and Xiaoqian Wang.
\newblock Difficulty-{Based} {Sampling} for {Debiased} {Contrastive} {Representation} {Learning}.
\newblock In \emph{2023 {IEEE}/{CVF} {Conference} on {Computer} {Vision} and {Pattern} {Recognition} ({CVPR})}, pages 24039--24048, 2023.

\bibitem[Kabra et~al.(2023)Kabra, Lewis, and Balakrishnan]{kabra_gelda_2023}
Krish Kabra, Kathleen~M. Lewis, and Guha Balakrishnan.
\newblock {GELDA}: {A} generative language annotation framework to reveal visual biases in datasets, 2023.
\newblock arXiv:2311.18064.

\bibitem[Karras et~al.(2018)Karras, Aila, Laine, and Lehtinen]{karras_progressive_2018}
Tero Karras, Timo Aila, Samuli Laine, and Jaakko Lehtinen.
\newblock Progressive {Growing} of {GANs} for {Improved} {Quality}, {Stability}, and {Variation}, 2018.
\newblock arXiv:1710.10196.

\bibitem[Kim et~al.(2021)Kim, Lee, and Choo]{kim_biaswap_2021}
Eungyeup Kim, Jihyeon Lee, and Jaegul Choo.
\newblock {BiaSwap}: {Removing} {Dataset} {Bias} with {Bias}-{Tailored} {Swapping} {Augmentation}.
\newblock In \emph{2021 {IEEE}/{CVF} {International} {Conference} on {Computer} {Vision} ({ICCV})}, pages 14972--14981, 2021.

\bibitem[Kim et~al.(2024{\natexlab{a}})Kim, Hwang, Ahn, Park, and Kwak]{kim_learning_2024}
Nayeong Kim, Sehyun Hwang, Sungsoo Ahn, Jaesik Park, and Suha Kwak.
\newblock Learning debiased classifier with biased committee.
\newblock In \emph{Proceedings of the 36th {International} {Conference} on {Neural} {Information} {Processing} {Systems}}, pages 18403--18415, 2024{\natexlab{a}}.

\bibitem[Kim et~al.(2024{\natexlab{b}})Kim, Mo, Kim, Lee, Lee, and Shin]{kim_discovering_2024}
Younghyun Kim, Sangwoo Mo, Minkyu Kim, Kyungmin Lee, Jaeho Lee, and Jinwoo Shin.
\newblock Discovering and {Mitigating} {Visual} {Biases} through {Keyword} {Explanation}, 2024{\natexlab{b}}.
\newblock arXiv:2301.11104.

\bibitem[Kingma and Ba(2017)]{kingma_adam_2017}
Diederik~P. Kingma and Jimmy Ba.
\newblock Adam: {A} {Method} for {Stochastic} {Optimization}, 2017.
\newblock arXiv:1412.6980.

\bibitem[Krishnakumar et~al.(2021)Krishnakumar, Prabhu, Sudhakar, and Hoffman]{krishnakumar_udis_2021}
Arvind Krishnakumar, Viraj Prabhu, Sruthi Sudhakar, and Judy Hoffman.
\newblock {UDIS}: {Unsupervised} {Discovery} of {Bias} in {Deep} {Visual} {Recognition} {Models}.
\newblock In \emph{British {Machine} {Vision} {Conference} ({BMVC})}, 2021.

\bibitem[Lee et~al.(2020)Lee, Liu, Wu, and Luo]{lee_maskgan_2020}
Cheng-Han Lee, Ziwei Liu, Lingyun Wu, and Ping Luo.
\newblock {MaskGAN}: {Towards} {Diverse} and {Interactive} {Facial} {Image} {Manipulation}, 2020.
\newblock arXiv:1907.11922.

\bibitem[Lee et~al.(2022)Lee, Hoffman, Wang, and Chau]{lee_viscuit_2022}
Seongmin Lee, Judy Hoffman, Zijie~J. Wang, and Duen~Horng Chau.
\newblock {VIsCUIT}: {Visual} {Auditor} for {Bias} in {CNN} {Image} {Classifier}.
\newblock In \emph{2022 {IEEE}/{CVF} {Conference} on {Computer} {Vision} and {Pattern} {Recognition} ({CVPR})}, pages 21443--21451, 2022.

\bibitem[Li et~al.(2018)Li, Wu, Peng, Ernst, and Fu]{li_tell_2018}
Kunpeng Li, Ziyan Wu, Kuan-Chuan Peng, Jan Ernst, and Yun Fu.
\newblock Tell {Me} {Where} to {Look}: {Guided} {Attention} {Inference} {Network}.
\newblock In \emph{Proceedings of the {IEEE}/{CVF} {Conference} on {Computer} {Vision} and {Pattern} {Recognition} ({CVPR})}, pages 9215--9223, 2018.

\bibitem[Li et~al.(2022{\natexlab{a}})Li, Hoogs, and Xu]{li2022discover}
Zhiheng Li, Anthony Hoogs, and Chenliang Xu.
\newblock Discover and {Mitigate} {Unknown} {Biases} with {Debiasing} {Alternate} {Networks}.
\newblock In \emph{Computer {Vision} – {ECCV} 2022}, pages 270--288, 2022{\natexlab{a}}.

\bibitem[Li et~al.(2022{\natexlab{b}})Li, Hoogs, and Xu]{li_discover_2022}
Zhiheng Li, Anthony Hoogs, and Chenliang Xu.
\newblock Discover and {Mitigate} {Unknown} {Biases} with {Debiasing} {Alternate} {Networks}, 2022{\natexlab{b}}.
\newblock arXiv:2207.10077.

\bibitem[Liang et~al.(2023)Liang, Perona, and Balakrishnan]{liang_benchmarking_2023}
Hao Liang, Pietro Perona, and Guha Balakrishnan.
\newblock Benchmarking {Algorithmic} {Bias} in {Face} {Recognition}: {An} {Experimental} {Approach} {Using} {Synthetic} {Faces} and {Human} {Evaluation}.
\newblock In \emph{Proceedings of the IEEE/CVF International Conference on Computer Vision (ICCV)}, pages 4977--4987, 2023.

\bibitem[Lingenfelter et~al.(2022)Lingenfelter, Davis, and Hand]{bebis_quantitative_2022}
Bryson Lingenfelter, Sara~R. Davis, and Emily~M. Hand.
\newblock A {Quantitative} {Analysis} of {Labeling} {Issues} in the {CelebA} {Dataset}.
\newblock In \emph{Advances in {Visual} {Computing}}, pages 129--141. Springer International Publishing, Cham, 2022.

\bibitem[Liu et~al.(2019)Liu, Ding, Xia, Liu, Ding, Zuo, and Wen]{liu_stgan_2019}
Ming Liu, Yukang Ding, Min Xia, Xiao Liu, Errui Ding, Wangmeng Zuo, and Shilei Wen.
\newblock {STGAN}: {A} {Unified} {Selective} {Transfer} {Network} for {Arbitrary} {Image} {Attribute} {Editing}.
\newblock In \emph{2019 {IEEE}/{CVF} {Conference} on {Computer} {Vision} and {Pattern} {Recognition} ({CVPR})}, pages 3668--3677, 2019.

\bibitem[Liu et~al.(2015)Liu, Luo, Wang, and Tang]{liu_deep_2015}
Ziwei Liu, Ping Luo, Xiaogang Wang, and Xiaoou Tang.
\newblock Deep {Learning} {Face} {Attributes} in the {Wild}.
\newblock In \emph{2015 {IEEE} {International} {Conference} on {Computer} {Vision} ({ICCV})}, pages 3730--3738, 2015.

\bibitem[Luccioni et~al.(2023)Luccioni, Akiki, Mitchell, and Jernite]{luccioni_stable_2023}
Alexandra~Sasha Luccioni, Christopher Akiki, Margaret Mitchell, and Yacine Jernite.
\newblock Stable {Bias}: {Analyzing} {Societal} {Representations} in {Diffusion} {Models}, 2023.
\newblock arXiv:2303.11408.

\bibitem[Mehrabi et~al.(2022)Mehrabi, Morstatter, Saxena, Lerman, and Galstyan]{mehrabi_survey_2022}
Ninareh Mehrabi, Fred Morstatter, Nripsuta Saxena, Kristina Lerman, and Aram Galstyan.
\newblock A {Survey} on {Bias} and {Fairness} in {Machine} {Learning}.
\newblock \emph{ACM Computing Surveys}, 54\penalty0 (6):\penalty0 1--35, 2022.

\bibitem[Meister et~al.(2023)Meister, Zhao, Wang, Ramaswamy, Fong, and Russakovsky]{meister_gender_2023}
Nicole Meister, Dora Zhao, Angelina Wang, Vikram~V. Ramaswamy, Ruth Fong, and Olga Russakovsky.
\newblock Gender {Artifacts} in {Visual} {Datasets}.
\newblock In \emph{2023 {IEEE} {International} {Conference} on {Computer} {Vision} ({ICCV})}, pages 4837--4848, 2023.

\bibitem[Nam et~al.(2020)Nam, Cha, Ahn, Lee, and Shin]{nam_learning_2020}
Junhyun Nam, Hyuntak Cha, Sungsoo Ahn, Jaeho Lee, and Jinwoo Shin.
\newblock Learning from {Failure}: {De}-biasing {Classifier} from {Biased} {Classifier}.
\newblock In \emph{Advances in {Neural} {Information} {Processing} {Systems}}, pages 20673--20684. 2020.

\bibitem[Pahl et~al.(2022)Pahl, Rieger, Möller, Wittenberg, and Schmid]{pahl_female_2022}
Jaspar Pahl, Ines Rieger, Anna Möller, Thomas Wittenberg, and Ute Schmid.
\newblock Female, white, 27? {Bias} {Evaluation} on {Data} and {Algorithms} for {Affect} {Recognition} in {Faces}.
\newblock In \emph{Proceedings of the 2022 {ACM} {Conference} on {Fairness}, {Accountability}, and {Transparency}}, pages 973--987, 2022.

\bibitem[Parraga et~al.(2023)Parraga, More, Oliveira, Gavenski, Kupssinskü, Medronha, Moura, Simões, and Barros]{parraga_fairness_2023}
Otavio Parraga, Martin~D. More, Christian~M. Oliveira, Nathan~S. Gavenski, Lucas~S. Kupssinskü, Adilson Medronha, Luis~V. Moura, Gabriel~S. Simões, and Rodrigo~C. Barros.
\newblock Fairness in {Deep} {Learning}: {A} {Survey} on {Vision} and {Language} {Research}.
\newblock \emph{ACM Computing Surveys}, page 3637549, 2023.

\bibitem[Pessach and Shmueli(2023)]{pessach_review_2023}
Dana Pessach and Erez Shmueli.
\newblock A {Review} on {Fairness} in {Machine} {Learning}.
\newblock \emph{ACM Computing Surveys}, 55\penalty0 (3):\penalty0 1--44, 2023.

\bibitem[Petsiuk et~al.(2018)Petsiuk, Das, and Saenko]{petsiuk2018rise}
Vitali Petsiuk, Abir Das, and Kate Saenko.
\newblock {RISE}: Randomized input sampling for explanation of black-box models.
\newblock In \emph{Proceedings of the British Machine Vision Conference (BMVC)}, 2018.

\bibitem[Pillai et~al.(2022)Pillai, Koohpayegani, Ouligian, Fong, and Pirsiavash]{pillai_consistent_2022}
Vipin Pillai, Soroush~Abbasi Koohpayegani, Ashley Ouligian, Dennis Fong, and Hamed Pirsiavash.
\newblock Consistent {Explanations} by {Contrastive} {Learning}.
\newblock In \emph{2022 {IEEE}/{CVF} {Conference} on {Computer} {Vision} and {Pattern} {Recognition} ({CVPR})}, pages 10203--10212, 2022.

\bibitem[Qraitem et~al.(2023)Qraitem, Saenko, and Plummer]{qraitem_bias_2023}
Maan Qraitem, Kate Saenko, and Bryan~A. Plummer.
\newblock Bias {Mimicking}: {A} {Simple} {Sampling} {Approach} for {Bias} {Mitigation}.
\newblock In \emph{Proceedings of the {IEEE}/{CVF} {Conference} on {Computer} {Vision} and {Pattern} {Recognition} ({CVPR})}, pages 20311--20320, 2023.

\bibitem[Ramaswamy et~al.(2021)Ramaswamy, Kim, and Russakovsky]{ramaswamy_fair_2021}
Vikram~V. Ramaswamy, Sunnie S.~Y. Kim, and Olga Russakovsky.
\newblock Fair {Attribute} {Classification} {Through} {Latent} {Space} {De}-{Biasing}.
\newblock In \emph{Proceedings of the {IEEE}/{CVF} {Conference} on {Computer} {Vision} and {Pattern} {Recognition} ({CVPR})}, pages 9301--9310, 2021.

\bibitem[Rao et~al.(2023)Rao, Böhle, Parchami-Araghi, and Schiele]{rao_studying_2023}
Sukrut Rao, Moritz Böhle, Amin Parchami-Araghi, and Bernt Schiele.
\newblock Studying {How} to {Efficiently} and {Effectively} {Guide} {Models} with {Explanations}.
\newblock In \emph{2023 {IEEE}/{CVF} {International} {Conference} on {Computer} {Vision} ({ICCV})}, pages 1922--1933, 2023.

\bibitem[Ribeiro et~al.(2016)Ribeiro, Singh, and Guestrin]{ribeiro2016lime}
Marco~Tulio Ribeiro, Sameer Singh, and Carlos Guestrin.
\newblock "{Why} {Should} {I} {Trust} {You}?": Explaining the {Predictions} of {Any} {Classifier}.
\newblock In \emph{Proceedings of the 22nd ACM SIGKDD International Conference on Knowledge Discovery and Data Mining}, page 1135–1144, 2016.

\bibitem[Russakovsky et~al.(2015)Russakovsky, Deng, Su, Krause, Satheesh, Ma, Huang, Karpathy, Khosla, Bernstein, Berg, and Fei-Fei]{russakovsky_imagenet_2015}
Olga Russakovsky, Jia Deng, Hao Su, Jonathan Krause, Sanjeev Satheesh, Sean Ma, Zhiheng Huang, Andrej Karpathy, Aditya Khosla, Michael Bernstein, Alexander~C. Berg, and Li Fei-Fei.
\newblock {ImageNet} {Large} {Scale} {Visual} {Recognition} {Challenge}.
\newblock \emph{International Journal of Computer Vision}, 115\penalty0 (3):\penalty0 211--252, 2015.

\bibitem[Sagawa et~al.(2020)Sagawa, Koh, Hashimoto, and Liang]{sagawa_distributionally_2020}
Shiori Sagawa, Pang~Wei Koh, Tatsunori~B. Hashimoto, and Percy Liang.
\newblock Distributionally {Robust} {Neural} {Networks}.
\newblock In \emph{International {Conference} on {Learning} {Representations}}, 2020.

\bibitem[Selvaraju et~al.(2020)Selvaraju, Cogswell, Das, Vedantam, Parikh, and Batra]{selvaraju_grad-cam_2020}
Ramprasaath~R. Selvaraju, Michael Cogswell, Abhishek Das, Ramakrishna Vedantam, Devi Parikh, and Dhruv Batra.
\newblock Grad-{CAM}: {Visual} {Explanations} from {Deep} {Networks} via {Gradient}-based {Localization}.
\newblock \emph{International Journal of Computer Vision}, 128\penalty0 (2):\penalty0 336--359, 2020.

\bibitem[Seo et~al.(2022)Seo, Lee, and Han]{seo_unsupervised_2022}
Seonguk Seo, Joon-Young Lee, and Bohyung Han.
\newblock Unsupervised {Learning} of {Debiased} {Representations} with {Pseudo}-{Attributes}.
\newblock In \emph{2022 {IEEE}/{CVF} {Conference} on {Computer} {Vision} and {Pattern} {Recognition} ({CVPR})}, pages 16721--16730, 2022.

\bibitem[Seshadri et~al.(2024)Seshadri, Singh, and Elazar]{seshadri_bias_2023}
Preethi Seshadri, Sameer Singh, and Yanai Elazar.
\newblock The {Bias} {Amplification} {Paradox} in {Text}-to-{Image} {Generation}.
\newblock In \emph{Proceedings of the 2024 {Conference} of the {North} {American} {Chapter} of the {Association} for {Computational} {Linguistics}: {Human} {Language} {Technologies} ({Volume} 1: {Long} {Papers})}, pages 6367--6384, 2024.

\bibitem[Shankar et~al.(2017)Shankar, Halpern, Breck, Atwood, Wilson, and Sculley]{shankar_no_2017}
Shreya Shankar, Yoni Halpern, Eric Breck, James Atwood, Jimbo Wilson, and D. Sculley.
\newblock No {Classification} without {Representation}: {Assessing} {Geodiversity} {Issues} in {Open} {Data} {Sets} for the {Developing} {World}, 2017.
\newblock arXiv:1711.08536.

\bibitem[Shitole et~al.(2021)Shitole, Li, Kahng, Tadepalli, and Fern]{shitole2021sag}
Vivswan Shitole, Fuxin Li, Minsuk Kahng, Prasad Tadepalli, and Alan Fern.
\newblock One explanation is not enough: Structured attention graphs for image classification.
\newblock In \emph{Advances in Neural Information Processing Systems}, pages 11352--11363. 2021.

\bibitem[Simonyan et~al.(2014)Simonyan, Vedaldi, and Zisserman]{simonyan_deep_2014}
Karen Simonyan, Andrea Vedaldi, and Andrew Zisserman.
\newblock Deep {Inside} {Convolutional} {Networks}: {Visualising} {Image} {Classification} {Models} and {Saliency} {Maps}, 2014.
\newblock arXiv:1312.6034.

\bibitem[Singh et~al.(2020)Singh, Mahajan, Grauman, Lee, Feiszli, and Ghadiyaram]{singh_dont_2020}
Krishna~Kumar Singh, Dhruv Mahajan, Kristen Grauman, Yong~Jae Lee, Matt Feiszli, and Deepti Ghadiyaram.
\newblock Don't {Judge} an {Object} by {Its} {Context}: {Learning} to {Overcome} {Contextual} {Bias}.
\newblock In \emph{Proceedings of the {IEEE}/{CVF} {Conference} on {Computer} {Vision} and {Pattern} {Recognition} ({CVPR})}, 2020.

\bibitem[Singla et~al.(2021)Singla, Nushi, Shah, Kamar, and Horvitz]{singla_understanding_2021}
Sahil Singla, Besmira Nushi, Shital Shah, Ece Kamar, and Eric Horvitz.
\newblock Understanding {Failures} of {Deep} {Networks} via {Robust} {Feature} {Extraction}.
\newblock In \emph{2021 {IEEE}/{CVF} {Conference} on {Computer} {Vision} and {Pattern} {Recognition} ({CVPR})}, pages 12848--12857, 2021.

\bibitem[Sohoni et~al.(2020)Sohoni, Dunnmon, Angus, Gu, and R{\'e}]{sohoni2020no}
Nimit Sohoni, Jared Dunnmon, Geoffrey Angus, Albert Gu, and Christopher R{\'e}.
\newblock No subclass left behind: Fine-grained robustness in coarse-grained classification problems.
\newblock \emph{Advances in Neural Information Processing Systems}, 33:\penalty0 19339--19352, 2020.

\bibitem[Tan and Le(2021)]{tan_efficientnetv2_2021}
Mingxing Tan and Quoc Le.
\newblock {EfficientNetV2}: {Smaller} {Models} and {Faster} {Training}.
\newblock In \emph{Proceedings of the 38th {International} {Conference} on {Machine} {Learning}}, pages 10096--10106. 2021.

\bibitem[Tartaglione et~al.(2021)Tartaglione, Barbano, and Grangetto]{tartaglione_end_2021}
Enzo Tartaglione, Carlo~Alberto Barbano, and Marco Grangetto.
\newblock {EnD}: {Entangling} and {Disentangling} deep representations for bias correction.
\newblock In \emph{2021 {IEEE}/{CVF} {Conference} on {Computer} {Vision} and {Pattern} {Recognition} ({CVPR})}, pages 13503--13512, 2021.

\bibitem[Torralba and Efros(2011)]{torralba_unbiased_2011}
Antonio Torralba and Alexei~A. Efros.
\newblock Unbiased look at dataset bias.
\newblock In \emph{{CVPR} 2011}, pages 1521--1528, 2011.

\bibitem[van Miltenburg(2016)]{van_miltenburg_stereotyping_2016}
Emiel van Miltenburg.
\newblock Stereotyping and {Bias} in the {Flickr30K} {Dataset}, 2016.
\newblock arXiv:1605.06083.

\bibitem[Vandenhende et~al.(2022)Vandenhende, Mahajan, Radenovic, and Ghadiyaram]{vandenhende2022counterfactual}
Simon Vandenhende, Dhruv Mahajan, Filip Radenovic, and Deepti Ghadiyaram.
\newblock Making heads or tails: Towards semantically consistent visual counterfactuals.
\newblock In \emph{ECCV}, 2022.

\bibitem[Verma and Rubin(2018)]{verma_fairness_2018}
Sahil Verma and Julia Rubin.
\newblock Fairness definitions explained.
\newblock In \emph{Proceedings of the {International} {Workshop} on {Software} {Fairness}}, pages 1--7, 2018.

\bibitem[Wah et~al.(2011)Wah, Branson, Welinder, Perona, and Belongie]{wah_caltech-ucsd_2011}
C. Wah, S. Branson, P. Welinder, P. Perona, and S. Belongie.
\newblock The {Caltech}-{UCSD} {Birds}-200-2011 dataset.
\newblock Technical Report CNS-TR-2011-001, California Institute of Technology, 2011.

\bibitem[Wang et~al.(2022)Wang, Liu, Zhang, Kleiman, Kim, Zhao, Shirai, Narayanan, and Russakovsky]{wang_revise_2022}
Angelina Wang, Alexander Liu, Ryan Zhang, Anat Kleiman, Leslie Kim, Dora Zhao, Iroha Shirai, Arvind Narayanan, and Olga Russakovsky.
\newblock {REVISE}: {A} {Tool} for {Measuring} and {Mitigating} {Bias} in {Visual} {Datasets}.
\newblock \emph{International Journal of Computer Vision}, 130\penalty0 (7):\penalty0 1790--1810, 2022.

\bibitem[Wang et~al.(2021)Wang, Liu, and Wang]{wang_are_2021}
Jialu Wang, Yang Liu, and Xin Wang.
\newblock Are {Gender}-{Neutral} {Queries} {Really} {Gender}-{Neutral}? {Mitigating} {Gender} {Bias} in {Image} {Search}.
\newblock In \emph{Proceedings of the 2021 {Conference} on {Empirical} {Methods} in {Natural} {Language} {Processing}}, pages 1995--2008, 2021.

\bibitem[Wang and Vasconcelos(2020)]{wang2020scout}
Pei Wang and Nuno Vasconcelos.
\newblock {SCOUT}: Self-{Aware} {Discriminant} {Counterfactual} {Explanations}.
\newblock In \emph{2020 IEEE/CVF Conference on Computer Vision and Pattern Recognition (CVPR)}, pages 8978--8987, 2020.

\bibitem[Wang et~al.(2019)Wang, Zhao, Yatskar, Chang, and Ordonez]{wang_balanced_2019}
Tianlu Wang, Jieyu Zhao, Mark Yatskar, Kai-Wei Chang, and Vicente Ordonez.
\newblock Balanced {Datasets} {Are} {Not} {Enough}: {Estimating} and {Mitigating} {Gender} {Bias} in {Deep} {Image} {Representations}.
\newblock In \emph{2019 {IEEE} {International} {Conference} on {Computer} {Vision} ({ICCV})}, pages 5310--5319, 2019.

\bibitem[Wang et~al.(2020)Wang, Qinami, Karakozis, Genova, Nair, Hata, and Russakovsky]{wang_towards_2020}
Zeyu Wang, Klint Qinami, Ioannis~Christos Karakozis, Kyle Genova, Prem Nair, Kenji Hata, and Olga Russakovsky.
\newblock Towards {Fairness} in {Visual} {Recognition}: {Effective} {Strategies} for {Bias} {Mitigation}.
\newblock In \emph{2020 {IEEE}/{CVF} {Conference} on {Computer} {Vision} and {Pattern} {Recognition} ({CVPR})}, pages 8916--8925, 2020.

\bibitem[Wolfe and Caliskan(2022)]{wolfe_american_2022}
Robert Wolfe and Aylin Caliskan.
\newblock American == {White} in {Multimodal} {Language}-and-{Image} {AI}.
\newblock In \emph{Proceedings of the 2022 {AAAI}/{ACM} {Conference} on {AI}, {Ethics}, and {Society}}, pages 800--812, 2022.

\bibitem[Wolfe et~al.(2023)Wolfe, Yang, Howe, and Caliskan]{wolfe_contrastive_2023}
Robert Wolfe, Yiwei Yang, Bill Howe, and Aylin Caliskan.
\newblock Contrastive {Language}-{Vision} {AI} {Models} {Pretrained} on {Web}-{Scraped} {Multimodal} {Data} {Exhibit} {Sexual} {Objectification} {Bias}.
\newblock In \emph{Proceedings of the 2023 {ACM} {Conference} on {Fairness}, {Accountability}, and {Transparency}}, pages 1174--1185, 2023.

\bibitem[Xiao et~al.(2021)Xiao, Engstrom, Ilyas, and Madry]{xiao_noise_2021}
Kai~Yuanqing Xiao, Logan Engstrom, Andrew Ilyas, and Aleksander Madry.
\newblock Noise or {Signal}: {The} {Role} of {Image} {Backgrounds} in {Object} {Recognition}.
\newblock In \emph{International {Conference} on {Learning} {Representations}}, 2021.

\bibitem[Xu et~al.(2021{\natexlab{a}})Xu, Wang, Chen, Zhou, and Loy]{xu_positional_2021}
Rui Xu, Xintao Wang, Kai Chen, Bolei Zhou, and Chen~Change Loy.
\newblock Positional {Encoding} as {Spatial} {Inductive} {Bias} in {GANs}.
\newblock In \emph{2021 {IEEE}/{CVF} {Conference} on {Computer} {Vision} and {Pattern} {Recognition} ({CVPR})}, pages 13564--13573, 2021{\natexlab{a}}.

\bibitem[Xu et~al.(2020)Xu, White, Kalkan, and Gunes]{xu_investigating_2020}
Tian Xu, Jennifer White, Sinan Kalkan, and Hatice Gunes.
\newblock Investigating {Bias} and {Fairness} in {Facial} {Expression} {Recognition}.
\newblock In \emph{Computer {Vision} – {ECCV} 2020 {Workshops}}, pages 506--523, 2020.

\bibitem[Xu et~al.(2021{\natexlab{b}})Xu, ZHANG, Zhang, and Tao]{xu_vitae_2021}
Yufei Xu, Qiming ZHANG, Jing Zhang, and Dacheng Tao.
\newblock Vitae: Vision transformer advanced by exploring intrinsic inductive bias.
\newblock In \emph{Advances in Neural Information Processing Systems}, pages 28522--28535. 2021{\natexlab{b}}.

\bibitem[Zeiler and Fergus(2014)]{zeiler_visualizing_2014}
Matthew~D. Zeiler and Rob Fergus.
\newblock Visualizing and {Understanding} {Convolutional} {Networks}.
\newblock In \emph{Computer {Vision} – {ECCV} 2014}, pages 818--833, 2014.

\bibitem[Zeng et~al.(2019)Zeng, Fu, Chao, and Guo]{zeng_learning_2019}
Yanhong Zeng, Jianlong Fu, Hongyang Chao, and Baining Guo.
\newblock Learning {Pyramid}-{Context} {Encoder} {Network} for {High}-{Quality} {Image} {Inpainting}.
\newblock In \emph{2019 {IEEE}/{CVF} {Conference} on {Computer} {Vision} and {Pattern} {Recognition} ({CVPR})}, pages 1486--1494, 2019.

\bibitem[Zhang et~al.(2018)Zhang, Bargal, Lin, Brandt, Shen, and Sclaroff]{zhang_top-down_2016}
Jianming Zhang, Sarah~Adel Bargal, Zhe Lin, Jonathan Brandt, Xiaohui Shen, and Stan Sclaroff.
\newblock Top-{Down} {Neural} {Attention} by {Excitation} {Backprop}.
\newblock \emph{International Journal of Computer Vision}, 126\penalty0 (10):\penalty0 1084--1102, 2018.

\bibitem[Zhao et~al.(2021)Zhao, Wang, and Russakovsky]{zhao_understanding_2021}
Dora Zhao, Angelina Wang, and Olga Russakovsky.
\newblock Understanding and {Evaluating} {Racial} {Biases} in {Image} {Captioning}.
\newblock In \emph{Proceedings of the {IEEE}/{CVF} {International} {Conference} on {Computer} {Vision} ({ICCV})}, pages 14830--14840, 2021.

\bibitem[Zhao et~al.(2017)Zhao, Wang, Yatskar, Ordonez, and Chang]{zhao_men_2017}
Jieyu Zhao, Tianlu Wang, Mark Yatskar, Vicente Ordonez, and Kai-Wei Chang.
\newblock Men {Also} {Like} {Shopping}: {Reducing} {Gender} {Bias} {Amplification} using {Corpus}-level {Constraints}.
\newblock In \emph{Proceedings of the 2017 {Conference} on {Empirical} {Methods} in {Natural} {Language} {Processing}}, pages 2979--2989, 2017.

\bibitem[Zhou et~al.(2016)Zhou, Khosla, Lapedriza, Oliva, and Torralba]{zhou_learning_2016}
Bolei Zhou, Aditya Khosla, Agata Lapedriza, Aude Oliva, and Antonio Torralba.
\newblock Learning {Deep} {Features} for {Discriminative} {Localization}.
\newblock In \emph{2016 {IEEE} {Conference} on {Computer} {Vision} and {Pattern} {Recognition} ({CVPR})}, pages 2921--2929, 2016.

\bibitem[Zhou et~al.(2018)Zhou, Lapedriza, Khosla, Oliva, and Torralba]{zhou_places_2018}
Bolei Zhou, Agata Lapedriza, Aditya Khosla, Aude Oliva, and Antonio Torralba.
\newblock Places: {A} 10 {Million} {Image} {Database} for {Scene} {Recognition}.
\newblock \emph{IEEE Transactions on Pattern Analysis and Machine Intelligence}, 40\penalty0 (6):\penalty0 1452--1464, 2018.

\bibitem[Zhou et~al.(2021)Zhou, Zhang, Zhang, Zhang, Bao, Chen, Zhang, and Wen]{zhou_cocosnet_2021}
Xingran Zhou, Bo Zhang, Ting Zhang, Pan Zhang, Jianmin Bao, Dong Chen, Zhongfei Zhang, and Fang Wen.
\newblock {CoCosNet} v2: {Full}-{Resolution} {Correspondence} {Learning} for {Image} {Translation}.
\newblock In \emph{2021 Proceedings of the IEEE/CVF Conference on Computer Vision and Pattern Recognition (CVPR)}, pages 11465--11475, 2021.

\end{thebibliography}
}

\clearpage
\maketitlesupplementary
\appendix

\section{Gradients for GradCAM}
\label{sec:gradcam-details}

In \cref{sec:gradcam}, to compute GradCAM for image features that contribute positively, we describe taking the gradient of the absolute value of the class output $|y_a|$ for binary cross-entropy loss, while taking gradient of class output $y_a$ directly for categorical cross-entropy loss. 

When using a model that is trained using binary cross-entropy loss, computing the gradient \wrt the absolute value of the logit (before the sigmoid) is equivalent to computing the gradient \wrt to the predicted class for categorical cross-entropy loss with two heads (one each for the positive and negative class). Concretely, let $s$ be the value of the logit; the probability that this model assigns to the positive class is $\sigma(s) = \frac{1}{1 + e^{-s}}$, and the probability assigned to the negative class is $1-\sigma(s) = \frac{e^{-s}}{1 + e^{-s}} = \sigma(-s)$. The model prediction is $\arg \max(\sigma(s), \sigma(-s)) = \arg\max(s, -s)$. Thus, taking the gradient with respect to the absolute value of the logits allows us to find positive contributions to the predicted binary class.

\section{Proofs of Invariants}
\label{sec:proofs}

In \cref{sec:method-metric}, we introduce the \metric{} metric, $\calB_{\text{A-IoU}}$, which is invariant to scale and size for pixel maps.

First, we confirm that if the two input maps are identical, $\bM_1 = \bM_2 = \bM \in \reals^{h\times w}$, the \metric{} metric is 1:
\begin{align}
    \calB_{\text{A-IoU}}(\bM, \bM) &= \dfrac{\cyc{\widehat{\bM}, \widehat{\bM}}_F}{\left\|\frac{\widehat{\bM}+\widehat{\bM}}{2}\right\|_F^2} \\
    &= \dfrac{\cyc{\widehat{\bM}, \widehat{\bM}}_F}{\left\|\widehat{\bM}\right\|_F^2} = \dfrac{\left\|\widehat{\bM}\right\|_F^2}{\left\|\widehat{\bM}\right\|_F^2} = 1.
\end{align}

We next prove that $\calB_{\text{A-IoU}}$ is scale invariant. Given two maps $\bM_1, \bM_2 \in \reals^{h\times w}$, suppose the maps are multiplied by the scalars $a_1, a_2 \in \reals_+$ respectively. Then their $L_1$ normalized maps are 
\begin{align}
    \widehat{a_i\bM}_i = \frac{a_i\bM_i}{||a_i\bM_i||_1} = \frac{a_i\bM_i}{a_i||\bM_i||_1} = \widehat{\bM}_i
\end{align}
So $\calB_{\text{A-IoU}}(a_1\bM_1, a_2\bM_2) = \calB_{\text{A-IoU}}(\bM_1, \bM_2)$.

For the proof of size invariance, we assume for simplicity that the maps are resized by a positive integer scalar $\alpha \in \nats$ using nearest neighbor interpolation. Again, consider two maps $\bM_1, \bM_2 \in \reals^{h\times w}$. Let $\bM^\alpha_1, \bM^\alpha_2 \in \reals^{\alpha h\times \alpha w}$ be the rescaling of the two maps by the constant $\alpha$. For example, with $\alpha=2$, a $5\times5$ box in the center of the map will be resized to be a $10\times10$ box, with the same spacial location within the map. Note that the $L_1$ normalized maps are
\begin{align}
    \widehat{\bM}^\alpha_i = \frac{\bM^\alpha_i}{||\bM^\alpha_i||_1} = \frac{\bM^\alpha_i}{\alpha^2||\bM_i||_1},
\end{align}
as each pixel in the original map appears $\alpha^2$ times in the resized map. Furthermore, the Frobenius inner product of the two resized maps is
\begin{align}
    \cyc{\bM^\alpha_1, \bM^\alpha_2}_F &= \sum_{i=1}^{\alpha h}\sum_{j=1}^{\alpha w} (\bM^\alpha_1)_{ij} \cdot (\bM^\alpha_2)_{ij} \\
    &= \alpha^2\sum_{i=1}^h\sum_{j=1}^w (\bM_1)_{ij} \cdot (\bM_2)_{ij} \\
    &= \alpha^2\cyc{\bM_1, \bM_2}_F
\end{align}
and, for the norm,
\begin{align}
    \left\|\frac{\widehat{\bM}^\alpha_1+\widehat{\bM}^\alpha_2}{2}\right\|_F^2 &= \frac{1}{4}\sum_{i=1}^{\alpha h}\sum_{j=1}^{\alpha w}\left(\frac{(\bM^\alpha_1)_{ij}}{||\bM^\alpha_1||_1} + \frac{(\bM^\alpha_2)_{ij}}{||\bM^\alpha_2||_1}\right)^2 \\
    &= \frac{1}{4\alpha^4}\sum_{i=1}^{\alpha h}\sum_{j=1}^{\alpha w}\left(\frac{(\bM^\alpha_1)_{ij}}{||\bM_1||_1} + \frac{(\bM^\alpha_2)_{ij}}{||\bM_2||_1}\right)^2 \\
    &= \frac{1}{4\alpha^2}\sum_{i=1}^{h}\sum_{j=1}^{w}\left(\frac{(\bM_1)_{ij}}{||\bM_1||_1} + \frac{(\bM_2)_{ij}}{||\bM_2||_1}\right)^2 \\
    &= \frac{1}{\alpha^2}\left\|\frac{\widehat{\bM}_1+\widehat{\bM}_2}{2}\right\|_F^2.
\end{align}
Thus, combining the two parts together,
\begin{align}
    \calB_{\text{A-IoU}}(\bM^\alpha_1, \bM^\alpha_2) &= \dfrac{\cyc{\widehat{\bM}^\alpha_1, \widehat{\bM}^\alpha_2}_F}{\left\|\frac{\widehat{\bM}^\alpha_1+\widehat{\bM}^\alpha_2}{2}\right\|_F^2} \\
    &= \dfrac{\frac{1}{\alpha^4||\bM_1||_1\cdot||\bM_2||_1}\cyc{\bM^\alpha_1, \bM^\alpha_2}_F}{\frac{1}{\alpha^2}\left\|\frac{\widehat{\bM}_1+\widehat{\bM}_2}{2}\right\|_F^2} \\
    &= \dfrac{\frac{1}{\alpha^2}\frac{1}{||\bM_1||_1\cdot||\bM_2||_1}\cyc{\bM_1, \bM_2}_F}{\frac{1}{\alpha^2}\left\|\frac{\widehat{\bM}_1+\widehat{\bM}_2}{2}\right\|_F^2} \\
    &= \dfrac{\cyc{\widehat{\bM}_1, \widehat{\bM}_2}_F}{\left\|\frac{\widehat{\bM}_1+\widehat{\bM}_2}{2}\right\|_F^2} \\
    &= \calB_{\text{A-IoU}}(\bM_1, \bM_2).
\end{align}
Although in the proof $\bM^\alpha_1$ and $\bM^\alpha_2$ are larger matrices than $\bM_1$ and $\bM_2$, the same argument applies if $\bM_1$ and $\bM_2$ are zero-padded to have same dimensions as the resized maps.

\section{Subsampling Training Details}
\label{sec:subgroup-details}

\begin{figure}[h]
    \centering
    \begin{subfigure}{0.85\linewidth}
        \centering
        \includegraphics[width=\linewidth]{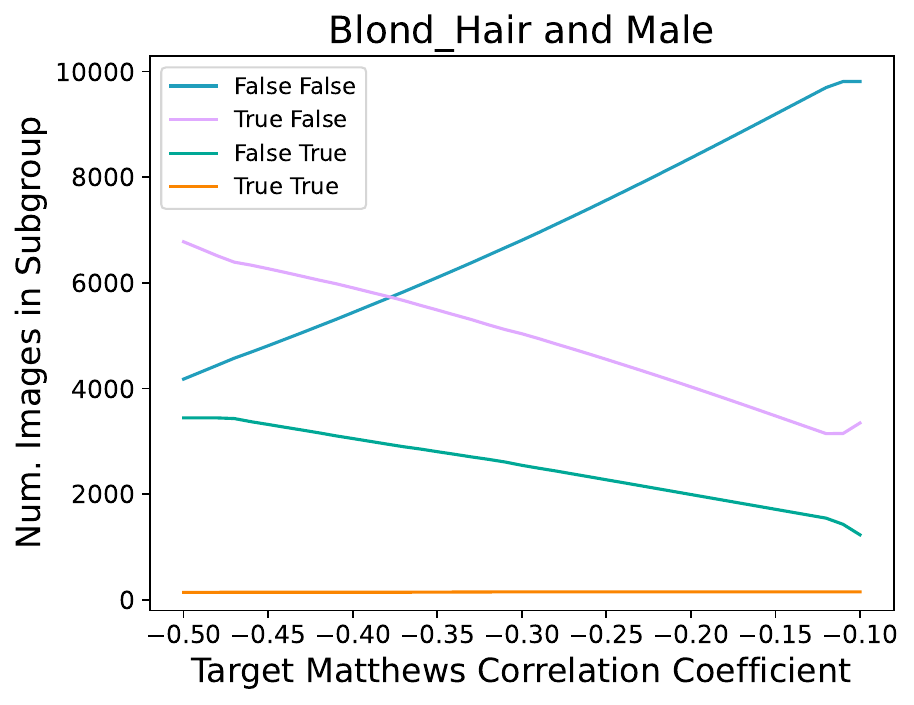}
    \end{subfigure}\\
    \begin{subfigure}{0.85\linewidth}
        \centering
        \includegraphics[width=\linewidth]{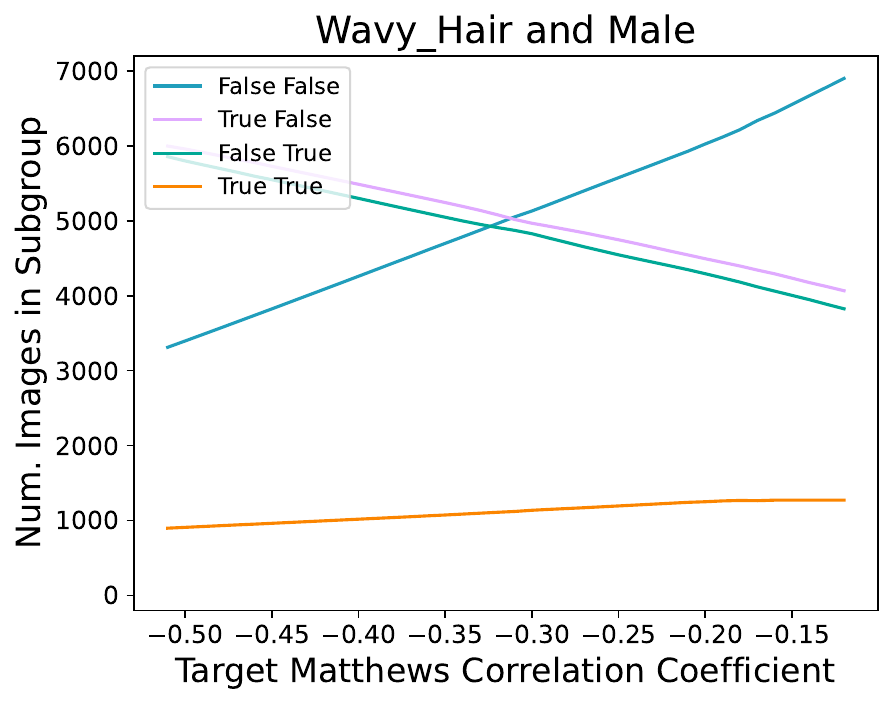}
    \end{subfigure}
    \caption{\textbf{Training set subgroup sizes under subsampling}. Here we report subgroup sizes of the training set of varying MCCs for \texttt{Blond\_Hair} and \texttt{Wavy\_Hair} with \texttt{Male}, under our optimization scheme, to compute the results in \cref{subsec:male-comparison} and \cref{fig:subgroup}. Subgroup sizes are bounded to the smallest subsampled training set size. The legend shows the four different subgroups groups, with the first value indicating the target label and the second \texttt{Male}.}
    \label{fig:subgroup_sampling}
\end{figure}

Here we provide experimental details for varying training set correlations in \cref{subsec:male-comparison}. Given a target Matthews correlation coefficient between the specified attribute and \texttt{Male}, we find subgroup sizes that achieve the target MCC (as MCC is dependent entirely on the sizes of the 4 subgroups) using SciPy's \texttt{optimize.minimize} with the trust region method\footnote{\url{https://docs.scipy.org/doc/scipy/reference/optimize.minimize-trustconstr.html}} (\cref{fig:subgroup_sampling}). We bound the sizes of the subsampled subgroups to the size of the original groups, and aim to minimize the distance to the original group sizes by the $L_2$ norm. To reduce fluctuations between the subsampled sizes, we initialize the optimizer with the adjacent subgroup sizes, with the original subgroups sizes in the training set as the starting point. Lastly, after running the optimization once for all MCCs, we rerun the optimization process with the additional bound of the smallest subsampled training set, so that all the subsampled training sets are of the same size. As the subsampling was an ablation study, the heatmap scores reported in \cref{fig:subgroup} were run on the validation set.

\section{Additional CelebA Results}
\label{sec:celeb-additional}

\begin{figure}[h]
    \centering
    \setlength{\tabcolsep}{2pt}
    \small{\begin{tabularx}{0.9\linewidth} { 
          >{\centering\arraybackslash}X  >{\centering\arraybackslash}X 
          >{\centering\arraybackslash}X  
          >{\centering\arraybackslash}X}
        \includegraphics[width=0.85\linewidth]{figures/cam_male.png} &
        \includegraphics[width=0.85\linewidth]{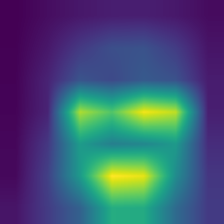} &
        \includegraphics[width=0.85\linewidth]{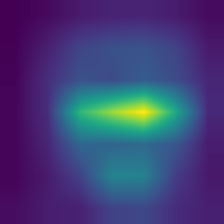} \\        \texttt{Male} & \texttt{Wearing\_} \texttt{Lipstick} & \texttt{Eyeglasses} \\
        \includegraphics[width=0.85\linewidth]{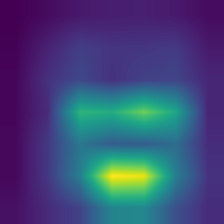} &
        \includegraphics[width=0.85\linewidth]{figures/cam_blond_hair.png} & 
        \includegraphics[width=0.85\linewidth]{figures/cam_wavy_hair.png} \\
        \texttt{Mustache} & \texttt{Blond\_Hair} & \texttt{Wavy\_Hair}
    \end{tabularx}}
    \caption{\textbf{Average heatmaps for CelebA attributes}. We visualize average heatmaps for the selected attributes investigated in \cref{subsec:male-comparison}.}
    \label{fig:celeba_heatmaps}
\end{figure}

\begin{figure}[ht]
    \centering
    \includegraphics[width=\linewidth]{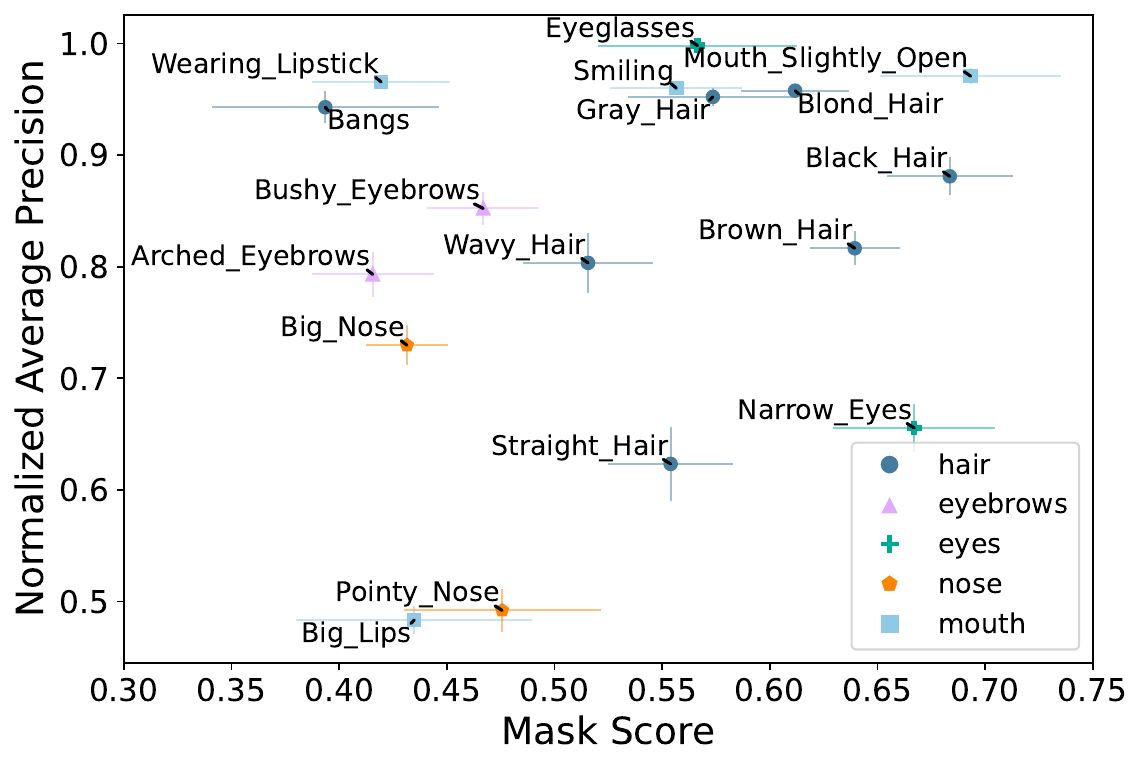}
    \caption{\textbf{Evaluation of mask score using GradCAM on CelebA test set with attribute-specific feature masks, compared to average precision}. To compare per-attribute AP between attributes, we adopt Hoiem \etal's normalized average precision (\apnorm{}) metric~\cite{hoiem_diagnosing_2012}.}
    \label{fig:mask_scatter_ap}
\end{figure}

\smallsec{Model Evaluation} The average precision weighted for all 40 attributes in CelebA, averaged across the 20 trained models with the experimental setup detailed in \cref{subsec:ground-truth}, is $0.902 \pm 0.025$. For reference, the normalized average precision (\apnorm{})~\cite{hoiem_diagnosing_2012} for the \texttt{Male} attribute is $0.994 \pm 0.003$, the second highest after \texttt{Eyeglasses} ($0.998 \pm 0.001$). In \cref{fig:celeba_heatmaps} we show average heatmaps for select attributes.

\smallsec{CelebA Normalized Average Precision} As a comparison to \cref{fig:mask_scatter}, which shows CelebA mask score against worst group accuracy, in \cref{fig:mask_scatter_ap} we show the mask score of the same 17 attributes to their normalized average precision (\apnorm{}). Compared with worst group accuracy, there is a no correlation for normalized average precision with respect to the mask score. Unlike worst group accuracy, to calculate normalized average precision one does not need to assume the correlated attribute.

\section{Evaluating with EfficientNet}
\label{sec:efficientnet}

To demonstrate the effectiveness of \metric{} on architectures other than ResNet, we also evaluated the metric using the EfficientNetV2-S architecture~\cite{tan_efficientnetv2_2021} on both the Waterbirds and CelebA datasets. Aside from the change in architecture, and averaging over 10 trained models instead of 20, the experimental setup remained the same.

\begin{figure}[ht]
    \centering
    \includegraphics[width=\linewidth]{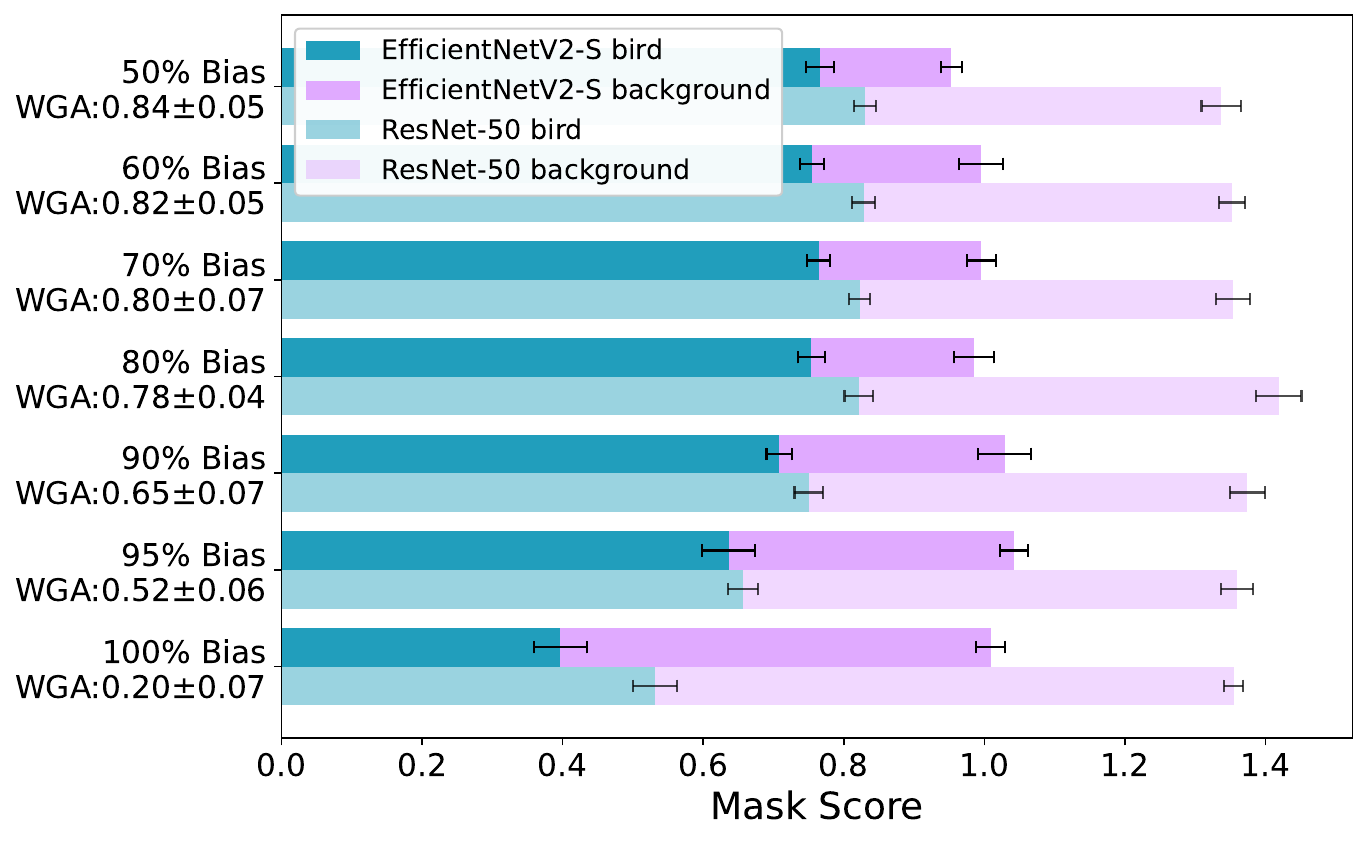}
    \centering
    \caption{\textbf{EfficientNetV2 mask score on Waterbirds}. 
    The top bars indicate \metric{} mask scores for EfficientNetV2-S models, while the bottom bars are corresponding ResNet-50 scores from Fig.~3.
    {WGA} is for the EfficientNet model. As with ResNet, the EfficientNet models attend less to the bird and more to the background, mirroring the decrease in WGA.
    }
    \label{fig:rebuttal-waterbirds}
\end{figure}

\begin{figure}[ht]
    \centering
    \includegraphics[width=\linewidth]{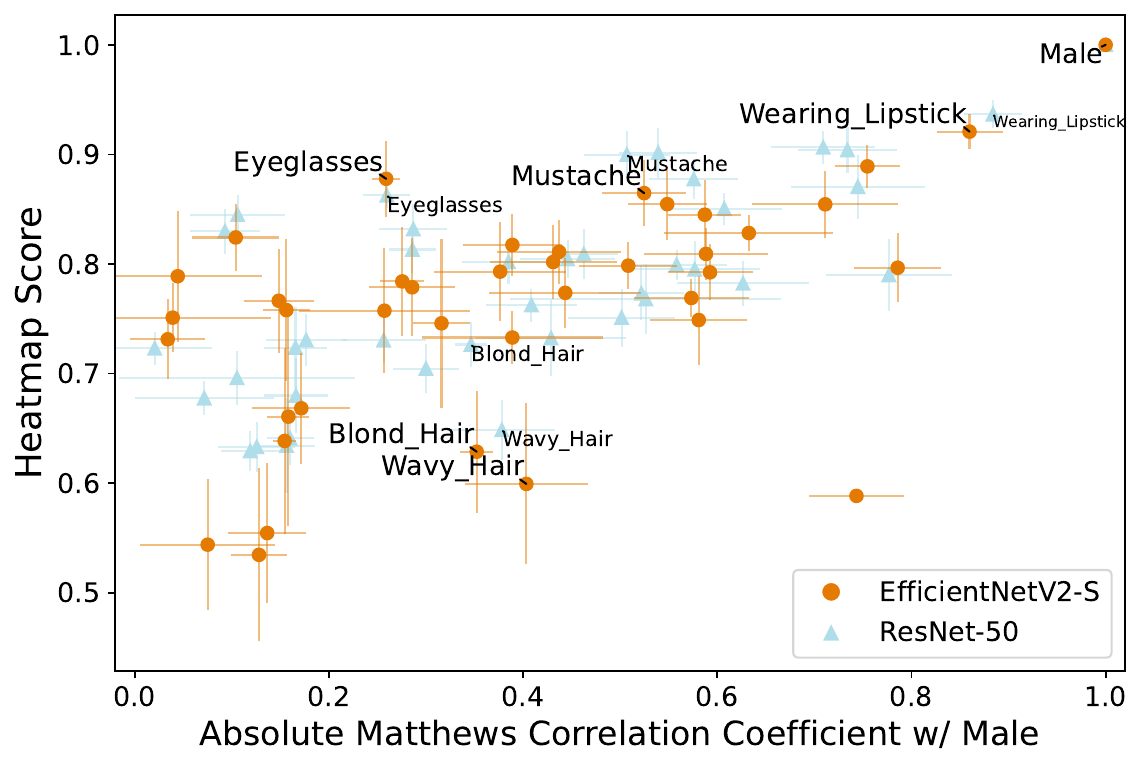}
    \centering
    \caption{\textbf{EfficientNetV2 heatmap scores on CelebA attributes.} Orange/circle indicates results with EfficientNetV2-S models, and light blue/triangle are ResNet-50 results from Fig.~5. We observe a very similar trend in EfficientNetV2 to that of ResNet-50. Highlighted attributes maintain their relative position, with some movement owing to different architectures and pretraining weights.}
    \label{fig:rebuttal-celeb}
    \vspace{-0.5cm}
\end{figure}

For Waterbirds, the EfficientNet models show a very similar pattern to ResNet in attending less to the bird and more to the background as dataset bias increases (\cref{fig:rebuttal-celeb}). The EfficientNet heatmap scores for CelebA also show a strong positive trend with MCC like ResNet (\cref{fig:rebuttal-waterbirds}). The 5 highlighted attributes maintain their relative positions, with some changes owing to different architectures and pretraining weights.

\vfill\eject

\end{document}